\documentclass[preprint,12pt]{elsarticle}

\usepackage{lineno,hyperref}
\modulolinenumbers[5]
\usepackage[table]{xcolor}
\usepackage{booktabs} 
\usepackage{hyperref}       
\usepackage{url}            
\usepackage{booktabs}       
\usepackage{amsfonts}       
\usepackage{nicefrac}       
\usepackage{microtype}      

\usepackage[utf8]{inputenc} 
\usepackage[english]{babel}
\usepackage[T1]{fontenc}    
\usepackage{cmap}
\usepackage{amsmath,amssymb,amsthm}             
\usepackage{dsfont}
\usepackage{mathtools}
\usepackage{amsthm}
\usepackage{amsmath}
\usepackage{bm}

\usepackage{algorithm}
\usepackage{algpseudocode}

\usepackage{paralist}
\usepackage{csquotes}
\usepackage{tkz-orm}

\usepackage{array}
\usepackage{multirow}
\usepackage{cleveref}

\usepackage{tikz}
\usetikzlibrary{babel}
\usetikzlibrary{decorations.pathmorphing} 
\usetikzlibrary{matrix} 
\usetikzlibrary{arrows} 
\usetikzlibrary{calc} 

\tikzset{add/.style n args={4}{
    minimum width=4mm,
    path picture={
        \draw[black, thick] 
            (path picture bounding box.south east) -- (path picture bounding box.north west)
            (path picture bounding box.south west) -- (path picture bounding box.north east);
        \node at ($(path picture bounding box.south)+(0,0.13)$)     {\tiny #1};
        \node at ($(path picture bounding box.west)+(0.13,0)$)      {\tiny #2};
        \node at ($(path picture bounding box.north)+(0,-0.13)$)        {\tiny #3};
        \node at ($(path picture bounding box.east)+(-0.13,0)$)     {\tiny #4};
        }
    }
}

\tikzstyle{block} = [draw,rectangle,thick,minimum height=2em,minimum width=2em]
\tikzstyle{sum} = [draw,circle,inner sep=0mm,minimum size=2mm]
\tikzstyle{connector} = [->,thick]
\tikzstyle{line} = [thick]
\tikzstyle{branch} = [circle,inner sep=0pt,minimum size=1mm,fill=black,draw=black]
\tikzstyle{guide} = []
\tikzstyle{snakeline} = [connector, decorate, decoration={pre length=0.2cm,
                         post length=0.2cm, snake, amplitude=.4mm,
                         segment length=2mm},thick, magenta, ->]

\renewcommand{\vec}[1]{\ensuremath{\boldsymbol{#1}}} 

\newcommand\cmbox[1]{
  \fbox{\lower0.75cm
    \vbox to 1.0cm{\vfil
      \hbox to 1.7cm{\hfil\parbox{1.4cm}{\centering #1}\hfil}
      \vfil}%
  }%
}

\newcommand\cmlegend[1]{
  {\lower0.75cm
    \vbox to 1.0cm{\vfil
      \hbox to 0.7cm{\hfil #1}
      \vfil}%
  }%
}

\DeclareMathOperator*{\argmin}{arg\,min}
\DeclareMathOperator*{\argmax}{arg\,max}

\def\automl{{\sc AutoML}}

\def\DPSH{{\sc DPSH}}
\def\CASH{{\sc CASH}}

\def\C3DE{{\sc }}
\def\SMAC{{\sc SMAC}}
\def\DAG{{\sc DAG}}
\def\AUTOSKLEARN{{\sc AutoSklearn}}
\def\AUTOSTACKER{{\sc AutoStacker}}
\def\MOSAIC{{\sc Mosaic}}
\def\TPOT{{\sc Tpot}}
\def\REINBO{{\sc ReinBo}}
\def\ALPHADM{{\sc Alpha3Dm}}

\def\SPEARMINT{{\sc Spearmint}}
\def\AUTOWEKA{{\sc AutoWeka}}
\def\HYPEROPT{{\sc Hyperopt}}
\def\HYPERBAND{{\sc Hyperband}}
\def\vsklearn{{\texttt sklearn}}
\def\vimblearn{{\texttt imblearn}}
\def\OpenML{{\sc OpenML-CC18}}
\usepackage[all]{hypcap}

\theoremstyle{definition}
\newtheorem{definition}{Definition}[section]

\journal{Information Systems}

\begin{document}

\begin{frontmatter}

\title{Two-stage Optimization for Machine Learning Workflow}

\author{Alexandre Quemy}
\address{IBM Krakow Software Lab, Cracow, Poland}
\ead{aquemy@pl.ibm.com}
\address{Faculty of Computing, Pozna\'n University of Technology, Pozna\'n, Poland}

\begin{abstract}
Machines learning techniques plays a preponderant role in dealing with massive amount of data and are employed in almost every possible domain. Building a high quality machine learning model to be deployed in production is a challenging task, from both, the subject matter experts and the machine learning practitioners. 

For a broader adoption and scalability of machine learning systems, the construction and configuration of machine learning workflow need to gain in automation. In the last few years, several techniques have been developed in this direction, known as \automl.

In this paper, we present a two-stage optimization process to build data pipelines and configure machine learning algorithms.
First, we study the impact of data pipelines compared to algorithm configuration in order to show the importance of data preprocessing over hyperparameter tuning. The second part presents policies to efficiently allocate search time between data pipeline construction and algorithm configuration. Those policies are agnostic from the metaoptimizer. Last, we present a metric to determine if a data pipeline is specific or independent from the algorithm, enabling fine-grain pipeline pruning and meta-learning for the coldstart problem.


\end{abstract}

\begin{keyword}
data pipelines, hyperparameter tuning, AutoML, CASH
\end{keyword}

\end{frontmatter}


\section{Introduction}

In practical machine learning, data are as important as algorithms. Algorithms received a lot of interest in hyperparameter tuning methods \cite{elshawi2019automated,hutter2019automatic}, that is to say, the art of adjusting parameters that are not dependent on the instance data. Contrarily, data pipeline construction and configuration received little if any interest. For instance, \cite{CRONE2006781} notices that algorithm hyperparameter tuning is performed in 16 out of 19 selected publications while only 2 publications study the impact of data preprocessing. This can probably be explained by the fact that the research community mainly uses ready-to-consume datasets, hence occulting de facto this problematic. In practice however, raw data are rarely ready to be consumed and must be transformed by a carefully selected sequence of preprocessing operations.

\begin{figure}[h!]
\includegraphics[width=1.10\textwidth,keepaspectratio]{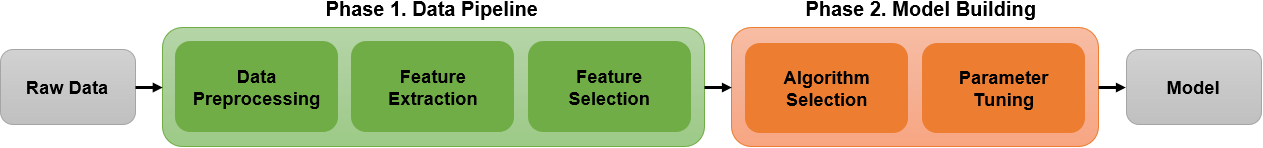}
\caption{Typical machine learning workflow}
\label{ml_workflow}
\end{figure}

In fact, building a high quality machine learning model to be deployed in production is a challenging task that is time consuming and computationally demanding. The usual machine learning workflow described by Figure \ref{ml_workflow} is broken down into two parts: 
\begin{enumerate}
\item finding the correct sequence of data transformations such that the dataset is consumable by a machine learning algorithm,
\item selecting the proper machine learning algorithm and its hyperparameters, such that the model provides a good generalization w.r.t. a given performance metric.
\end{enumerate}

On one hand, there are plenty of reasons that can explain why a data source cannot be used directly and require preprocessing: too many variables, imbalanced dataset, missing values, outliers, noise, specific domain restriction of the algorithms, etc. On the other hand, data preprocessing has a huge impact on the model performances \cite{dasu2003exploratory,CRONE2006781,NAWI201332}.


The data pipeline depends both on the data source and the algorithm such that there is no universal pipeline that can work for every data source and every algorithm \cite{wolpert1996lack}. The data pipeline is usually defined by trial and error, using the experience of data scientists and the expert knowledge about the data. This step is so crucial it may represent up to 80\% of data scientist time \cite{chessell2014governing}.

The techniques to automate the construction of machine learning workflows are called \automl.
Broadly speaking, \automl~consists in solving the following black-box optimization problem: given a dataset, 
\begin{equation}
\label{bbo}
\text{Find } \bm \lambda^* \in \underset{\bm \lambda \in \bm \Lambda}{\argmax}~ \mathcal F(\bm \lambda),
\end{equation} where $\bm \Lambda$ is the space of machine learning configuration, and $\mathcal F(\bm \lambda)$ the performance of the model learned over the dataset using the configuration $\bm \lambda$. 

This paper presents a two-stage optimization approach to solve the \automl~problem. Specifically, we are interested in understanding the respective impact of the data pipelines and the algorithm configurations on the model performances. Additionally, we study the best allocation of time between the two phases of the machine learning workflow.

To the best of our knowledge, most of \automl~systems tackles the problem by aggregating the data pipeline operators, the algorithm portfolio and their respective configuration space into a single gigantic search space (corresponding to $\bm \Lambda$ in Eq. \eqref{bbo}). In the approach presented in this paper, we divided the search for a solution into two steps, namely the data pipeline construction and configuration, and the algorithm selection and configuration.

Under the assumption that the two steps are roughly independent, the two-stage optimization brings the two following advantages:
\begin{enumerate}
\item by splitting the search space into two smaller ones, it speeds up the overall optimization process,
\item it is possible to statistically assess if a data pipeline is specific to an algorithm or rather {\it universal} w.r.t. the dataset, enabling meta-learning at a lower granularity level (see Section \ref{sec:exp_2}).
\end{enumerate}


The paper is an extension of the preliminary work \cite{DBLP:conf/dolap/Quemy19} that mainly focused on the influence of data pipelines on machine learning performances.

The rest of the paper is organized as follows: in Section \ref{sec:rw}, we present work related to \automl~ and discuss the limitations of current approaches. A reformulation of \CASH~ is proposed in Section \ref{sec:ml_workflow}. The two-stage optimization process is described in Section \ref{sec:two_stage}, followed by an experimental validation in Section \ref{sec:exp_1}. Finally, Section \ref{sec:exp_2} presents an indicator to evaluate the dependency of a data pipeline to the algorithm. 

\section{Related work}
\label{sec:rw}

In this section, we introduce the main problem the \automl~community focuses on, namely {\sc C}ombined {\sc A}lgorithm {\sc S}election and {\sc H}yperparameter-tuning (\CASH). After presenting the different approaches to solve it in Section \ref{sec:cash}, we discuss in Section \ref{sec:limit_cash}, why \CASH~is a difficult problem, why its formulation might not be the most suitable and why current tools are not satisfying w.r.t. to the way machine learning practitioners work.

For a broader review on \automl, we refer the reader to \cite{elshawi2019automated,hutter2019automatic}. 


\subsection{\CASH~problem}
\label{sec:cash}

The learning problem consists in finding or constructing an approximation of an unknown function $f \colon \mathcal X \to \mathcal Y$. 
A {\it learning} algorithm $A$ maps a set of training points $\{d_i\}_{i=1}^n$ with $d_i = (\mathbf x_i, y_i) \in \mathcal X \times \mathcal Y$ to $\mathcal{Y}^\mathcal{X}$.
A learning algorithm $A$ is parametrized by some hyperparameters $\bm \lambda \in \bm \Lambda$ that modify the way the algorithm $A_{\bm \lambda}$ learns. Each hyperparameter $\lambda_i$ belongs to a space $\Lambda_i$ and $\bm \Lambda$ is a subset of the cross-product of each domain, i.e. $\bm \Lambda \subset \Lambda_1 \times ... \times \Lambda_n$. In general, $\Lambda$ can be more structured (conditional tree, directed acycle graph,...).

\CASH~is formulated as the following optimization problem:

\begin{definition}[{\sc C}ombined {\sc A}lgorithm {\sc S}election and {\sc H}yperparameter-tuning]
  Given a portfolio of algorithms $\mathcal A = \{ A^{(1)},...,A^{(m)}\}$  with associated hyperparameter spaces $\bm \Lambda^{(1)},..., \bm \Lambda^{(m)}$, the Combined Algorithm Selection and Hyperparameter Optimization (\CASH) problem is defined by:
  \begin{displaymath}
    \label{eq:CASH}
    A^*_{\bm{\lambda}^*} = \underset{A^{(j)} \in \mathcal A, \bm{\lambda} \in \bm{\Lambda}^{(j)}}{\argmin} \frac 1 k \underset{i=1}{\overset{k}{\sum}} \mathcal L(A^{(j)}_{\bm \lambda}, \mathcal{D}^{(i)}_{\text{train}}, \mathcal{D}^{(i)}_{\text{test}}), \tag{CASH}
  \end{displaymath} where $\mathcal L$ is a loss function (e.g. error rate) obtained on the test set by the model learned by algorithm $A$ parametrized by $\bm \lambda$ over the training set.
\end{definition}

\subsection{Black-box optimization and surrogate learning}

The most basic technique for hyperparameter tuning is a grid search or factorial design \cite{montgomery2017design} which consists in exhaustively testing parameter configurations on a grid. In practice, this approach is computationally intractable. An efficient alternative, called Randomized search \cite{bergstra2012random}, consists in testing configurations (pseudo)randomly until a certain budget is exhausted.

Beyond those naive approaches, the most prominent approach to tackle \CASH~consists in iteratively building an approximation, called {\it surrogate model}, of the optimization objective.

Recall the black-box optimization problem (\ref{bbo}) and the optimization objective $\mathcal F$. At step $t$, surrogate model $\hat{\mathcal{F}}$ is learned from the history $\{\bm \lambda_k, \mathcal{F}(\bm \lambda_k) \}$ for $k=1...t$ of previously selected configurations and associated performances. An acquisition function is used to determined the most promising configuration for $\bm \lambda^*_{t+1}$.

The difference between the various surrogate approaches lies into the model space assumption and the acquisition function. 
Sequential Model-based Algorithm Configuration (\SMAC) \cite{Hutter:2011:SMO:2177360.2177404,Bergstra:2011:AHO:2986459.2986743} is based on Random Forest, so are frameworks based on \SMAC~such as \AUTOWEKA~\cite{thornton2013auto,kotthoff2017auto} or \AUTOSKLEARN~\cite{NIPS2015_5872}. \HYPEROPT~\cite{bergstra2015hyperopt} uses a Tree-structured Parzen Estimator (TPE) while \SPEARMINT~\cite{snoek2012practical} is based on Gaussian processes (GP).

An acquisition function is used to determined the next configuration to be sampled. Most of those functions are based on Bayesian optimization \cite{wilson2018maximizing,frazier2018tutorial}. One popular strategy is to select $\bm \lambda_{k+1}$ such that it maximizes the expected improvement \cite{movckus1975bayesian}.

As an alternative to Bayesian optimization, \cite{rakotoarison:hal-01966957} proposes to use Monte-Carlo Tree Search to iteratively explore a tree-structured search space while pruning the less promising configurations.

\subsection{Multi-fidelity optimization}

The black-box optimization problem (\ref{bbo}) is expensive: an iteration means preprocessing the whole dataset through the pipeline and then training a model, often many times as cross-validation is preferred to validate the result. Multi-fidelity optimization focuses on decreasing the computational cost by using large number of {\it cheap} low-fidelity evaluations. For this, several approaches have been considered. 

Extrapolating the model learning curve from the available training data provides an early stop criterion that reduces the computation time \cite{domhan2015speeding, swersky2014freeze}.

Bandit-based algorithms such as Successive halving \cite{jamieson2016non} or \HYPERBAND~\cite{li2018hyperband} find configurations by greedily allocating more budget to promising configurations. For instance, \HYPERBAND~randomly selects configurations and iteratively removes unpromising candidates while increasing the budget for the promising ones.

In \cite{DBLP:conf/bdas/NalepaMPHK18}, the authors uses a genetic algorithm to sample a subset of representative input vectors in order to speed-up the model training while increasing the model performances. Genetic algorithms are also used to search for the whole pipeline as in \TPOT \cite{olson2016evaluation} or \AUTOSTACKER \cite{chen2018autostacker}.

Last, despite a small configuration space, \REINBO \cite{DBLP:journals/corr/abs-1904-05381}, a reinforcement learning approach, has been shown to outperform TPE, \AUTOSKLEARN~and \TPOT~on many datasets.

\subsection{Meta-learning and warm-start}

Another research direction consist predicting and recommending good pipelines or operators for a given dataset or task. Referred to as meta-learning, it is particularly important for Bayesian optimization, extensively used to solve \CASH, since the quality of the results is conditioned by the surrogate initialization. To solve this problem, known as coldstart, several solutions have been investigated. 

In \AUTOSKLEARN~\cite{NIPS2015_5872}, about 140 datasets are represented as vectors made of 38 meta-features, and associated to the best pipeline ever found. When a new dataset is used, \AUTOSKLEARN~initializes the search process with the best configuration found for the closest dataset w.r.t. Euclidian distance in the meta-feature space. A more generic approach consists in learning the metric between datasets, using e.g. a Siamese Network~\cite{kim2017learning}.

At a lower level, \cite{Bilalli:2017:PPM:3214035.3214049} proposes to predict the impact of individual preprocessing operators rather than the whole pipeline.

For more extensive surveys on meta-learning, we refer the reader to \cite{vanschoren2018meta,elshawi2019automated}

\subsection{Limits of current approaches}
\label{sec:limit_cash}

The intrinsic difficulty of building a machine learning pipeline lies in the nature of the search space:
\begin{itemize}
\item the objective is non-separable i.e., the marginal performance of an operator $a$ depends on all the operators in all the paths leading to $a$ from the source,
\item within the configuration space of a specific operator $a$, there might be some dependencies between the hyperparameters (e.g. for Neural Networks, the coefficients $\alpha$, $\beta_1$ and $\beta_2$ make sense only for Adam solver \cite{kingma2014adam}).
\end{itemize}
Therefore, building a machine learning pipeline is a mix between selecting a proper sequence of operations and, for each operation, selecting the proper configuration in a structured and conditional space. On the contrary, most \automl~systems handle the problem by aggregating the whole search space, losing the sequential aspect of it. A notable exception is \MOSAIC~\cite{rakotoarison:hal-01966957}, inspired by \ALPHADM, that explores the search space in terms of actions on operators (insertion, deletion, etc.).


A second limitation is that most \automl~frameworks proposes a static search space and, even more constraining, a fixed {\it pipeline prototype} i.e., a high-level structure defining an ordered sequence of operator types (a precise definition is given in Section \ref{sec:ml_workflow}). For instance, \AUTOSKLEARN~ has a fixed pipeline made of one feature selection operator among 13 operators, and one up to three data preprocessing operator among only four. Those data preprocessing operators are of various nature: one-hot encoder, a specific imputation, balancing and rescaling method. There is no possibility to add custom operators nor to specify additional constraints in case some additional knowledge is available (e.g. no need for imputation since there is no missing values).

To the best of our knowledge, the only approach that uses a non-predetermined sequence of operators is \TPOT~\cite{olson2016evaluation}, but it is not possible to add additional constraints.


\section{Data Pipeline Selection and Hyperparameter Optimization}
\label{sec:ml_workflow}

\subsection{Operators and pipelines}

As mentioned before, most approaches model pipelines as fixed ordered sequences of $m$ algorithms and define for each step a specific set of algorithms that can be used. We define a more general version of pipelines, and then constrain this definition for a tradeoff between practical usage and flexibility.

We define a machine learning pipeline $\mathbf p$ as a Directed Acyclic Graph (\DAG) with two distinguished nodes: a {\it source}, from which all paths start, and a {\it sink}, from which all paths ends\footnote{The words {\it source} and {\it sink} come from Flow Network vocabulary.}. The source is the node by which the data enter, and the sink produces the solution to the machine learning problem. Generally, the nodes represent operators or algorithms. Note that the sink is not necessarily a machine learning algorithm but might also be an operator that aggregates the result of several algorithms (or paths) in an ensemble fashion.

Generally, an operator parametrized by $\bm \gamma \in \Gamma$ is a function defined by
\begin{align*}
  a_{\bm \gamma} \colon  \mathcal{X}_1^{n_1} &\to\mathcal{X}^{n_2}_2 \times \mathcal{X}_2^{\mathcal{X}_1}\\
  \mathcal D_1 &\mapsto \mathcal D_2, T_{\mathcal D_1}.
\end{align*}
The function operates over a dataset expressed in a certain representation space $\mathcal X_1$ and expresses it into another representation space $\mathcal X_2$. The size of the dataset may be modified (e.g. rebalancing operation), the input and output space may be the same (e.g. removing outliers) and the dimension of the input and output space may differ (e.g. Principal Component Analysis). The operator initially operates using the whole dataset (e.g. PCA or missing values imputation using a value derived from the available data such as the mean or median). It also returns a functor from $\mathcal{X}_1$ to $\mathcal{X}_2$ that is used to project a single element during the prediction phase. For instance, a PCA on the training set expresses this dataset in a $k$ dimensional space. The associated functor is the projector from the original space to the new space. For a rebalancing operator, as it operates only on the training set, the functor is the identity. Note that the functor is implicitly parametrized by $\bm \gamma$ (e.g. the parameter $k$ in a PCA is forwarded to the functor).

Two operators are {\bf compatible} if the representation output space $\mathcal X_2$ of the first operator is the same as the representation input space $\mathcal X_1$ of the second one. A pipeline is said {\bf compatible} if all the connected nodes are compatible. 

\subsection{\DPSH~problem}

We propose a reformulation of \CASH~that better takes into account the nature of machine learning pipelines and the way practitioners works.

Given a collection of operators $\mathcal A$, the Data Pipeline Selection and Hyperparameter Optimization (\DPSH) problem is defined by:
\begin{equation} \label{eq:dpsh}\tag{DPSH}
\mathbf{p}^*_{\bm{\gamma}^*} = \underset{m \in \mathbb{N}, \mathbf p \in \mathcal{G}_m(\mathcal A), \bm \gamma \in \bm{\Gamma}(\mathbf p)}{\argmin} \frac 1 k \underset{i=1}{\overset{k}{\sum}} \mathcal{L}(\mathbf{p}_{\bm\gamma}, \mathcal{D}^{(i)}_{\text{train}}, \mathcal{D}^{(i)}_{\text{test}}),
\end{equation} where $\mathcal{G}_m(\mathcal A)$ is the set of compatible pipelines with $m$ vertices selected in $\mathcal A$ (with possible replacement), and $\bm{\Gamma}(\mathbf p)$ the cartesian product of the configuration space of each operator in the pipeline $\mathbf p$. The size $m$ of the sequence is unknown a priori.



\subsection{Pipeline prototypes}

 Current approaches that intend to handle the pipeline construction uses a fixed sequence of few steps with a strict order, without possibility to change this topology. While this is useful for end-user with zero knowledge, it is also too restrictive for more complex workflows or end-user with slightly more advanced expertise. On the contrary, the problem depicted by \DPSH~is far too general to be directly handled because of the enormous number of operators one can imagine, and the number of possible compatible pipelines built over them.

\begin{figure}[h!]
\includegraphics[scale=0.4]{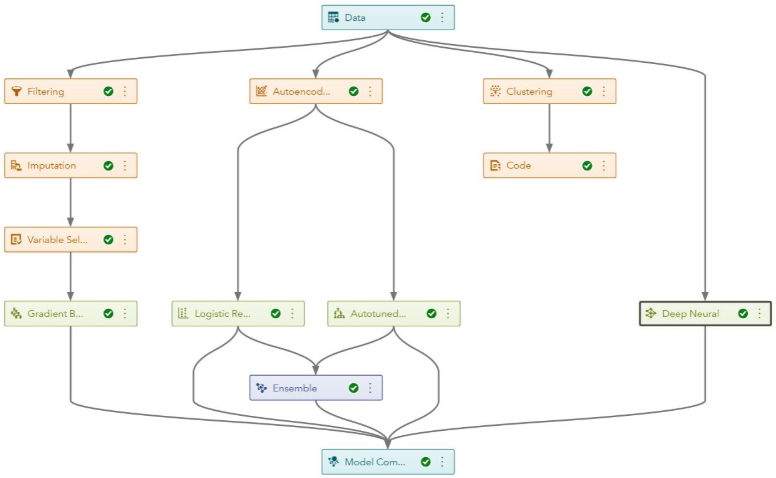}\hfill
\includegraphics[scale=0.4]{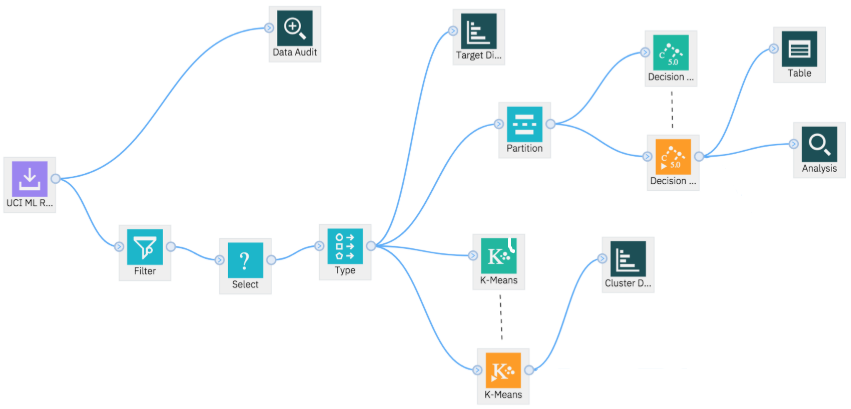}
\caption{Example of real-life pipelines designed with SAS (left) and IBM Watson Studio (right). The flow can be complex, and the topology totally depends on the problem to solve.}
\label{practical_workflow}
\end{figure}

Figure \ref{practical_workflow} depicts two real-life pipelines created by two different pipeline modelers. They are more complex than the usual small pipelines studied in the \automl~literature. Also, the network topology is highly different from one to another, notably because each pipeline is designed to answer a specific question on specific data. While the data scientist might not know exactly what concrete operations will perform the best, she has a general idea of the topology and some orders on the operations. This knowledge can be translated into constraints on the set of compatible pipelines.

We define a {\bf pipeline prototype} as a particular graph topology organized in {\bf layers}. Each layer groups specific operations by their purpose, with a level of granularity to be decided by the user. For instance, a layer could be specifically dedicated to imputation if the user knows some values are missing, or generically labelled as {\it feature engineering} with all possible sorts of preprocessing operators. An overly specialized prototype is easy to optimize but restricts the search space and thus the possibility to find outstanding configurations. On the contrary, too broadly defined prototype make the search very difficult as a lot of combinations are either non-feasible or lead to poor results. For instance, in Figure \ref{practical_workflow}, left part, a possible prototype could be made of four layers: data preprocessing (orange), model building (green), ensembling (dark blue) and model aggregation (light blue).

To control the pipeline compatibility, two combined approaches are used at the {\it software} level:
\begin{itemize}
  \item for each operator, we defined the input and output space in terms of types (e.g \texttt{int}, \texttt{float}, dictionary, vector of mixed types, etc.). When a pipeline is selected, a fast preliminary check can be done by graph traversal without having to feed the pipeline,
  \item during the execution, non-compatible pipelines throw an exception that can be caught as early as possible.
\end{itemize}
In both cases, the pipeline is declared {\it incompatible} by setting the loss function to $+\infty$, with the benefit of preventing the metaoptimizer to explore close areas of the search space, most likely incompatible as well.

Working with {\bf pipeline prototypes} allows to drastically reduce the search space of \DPSH~while aligning on real-life practices.

\section{Two-stage optimization under time-budget}
\label{sec:two_stage}

For our purpose, we consider a simplified version of the problem where the algorithm is given such that we are left with its configuration. To solve this problem, rather than considering one large search space, we break the optimization process into two smaller problems: searching for a good pipeline, and searching for a good algorithm configuration.

The rationale behind breaking down the \automl~optimization process lies into the distinct nature of the two steps described by Figure \ref{ml_workflow}. {\it Feature engineering} deals for most with improving the quality of the data, like a craftsman that transforms raw material into remarkable and consumable objects. In general, it has little to do with the person that will buy the object, or to speak less metaphorically, with the algorithm that consumes the preprocessed data. Obviously, this assumption is true up to a certain point since algorithms might have different structural requirements on their input space (e.g. categorical features for tree-based methods, and the influence of encoders on their performances), and some algorithms might be more sensitive to some preprocessing steps due to their mathematical properties. For instance, for a given time budget, a dimensionality reduction grants, theoretically a least, a better advantage to an algorithm with a $\mathcal~O(N^2)$ time complexity rather than a $\mathcal O(\log N)$ algorithm where $N$ is the initial input space dimension.

\subsection{Architecture}

\begin{figure}[h!]
\label{fig:architecture}
\hspace*{-1.8cm}
\begin{tikzpicture}[scale=1, auto, >=stealth']

    \small
    \matrix[ampersand replacement=\&, row sep=0.2cm, column sep=0.4cm] {
      \node[circle] (X1) {$\vec{X}$}; 
      \& \node[block] (D) {$\mathbf p, \vec \gamma_t$};
      \& \node[] (X) {$\vec{X}_t$};
      \& \node[block] (M1) {$A, \vec \lambda_{t,k}$}; 
      \& \node[] (Y) {$y_{t,k}$};
      \& \node[] (M2) {$(\vec \gamma, \vec \lambda)^*$};\\
       \& \& \&
      \node[block] (U1) {$\vec \lambda_{t,k+1} = f_A(\vec \lambda_{1:t,k}, \vec y_{1:t,k})$};\\
      \& \&
      \node[block] (U2) {$\vec \gamma_{t+1} = f_\mathbf p(\vec \gamma_{1:t}, \vec \lambda^*_{1:t}, \vec y^*_{1:t})$}; \&
      \node[] (Y2) {$( y^*_t, \vec \lambda^*_{t})$}; \\
    };

    \node[fit={(-0.05,-0.5) (5.7,2.4)}, inner sep=0pt, draw=black, thick] (rect) {};
    \node[fit={(-6.5,-1.5) (5.85,2.7)}, inner sep=0pt, draw=black, thick] (rect_2) {};

    \draw [connector] (X1) -- (D.west);

    \draw [connector] (D.east) -- (X);

    \draw [connector] (X) -- (M1);
    \draw [connector] (M1) -- (Y);
    \draw [connector] (Y) -- (M2);

    \draw [connector] (M1) -- (Y.west);
    \draw [connector] (rect.300) |- (Y2);
    \draw [connector] (Y2) -- (U2);
    \draw [connector] (Y.south) |- (U1);

    \draw [connector] (U1) -- (M1);
    \draw [connector] (U2) -| (D);

    \draw [magenta] ($(D.north) + (1.em, 2.7em)$) node {Pipeline phase} ;
    \draw [magenta] ($(M1.north) + (6.2em, 1.9em)$) node {Algo. phase} ;

    \draw [blue] ($(X1.south) - (2.5em, 0.1em)$) node {[full raw dataset]} ;
    \draw [blue] ($(M1.north) + (0em, 0.7em)$) node {[model]} ;
    \draw [blue] ($(D.north) + (0em, 0.7em)$) node {[pipeline]} ;

    \node[] (prior) at ($(M1.north) + (-3.2em, 1.8em)$) {$\lambda_{t+1,0} \gets \lambda^*_t$};
    \draw[blue] ($(M1.north) + (0.7em, 1.8em)$) node {[prior]} ;

    \draw [magenta] ($(X.north east) - (0em, 0.em)$) node {\ormind{1}} ;
    \draw [magenta] ($(rect.north east)  + (0.0em, 0.2em)$) node {\ormind{2}} ;
    \draw [magenta] ($(prior.north east)  + (0.2em, 0.2em)$) node {\ormind{3}} ;

    \draw [magenta] ($(Y2.north east) - (0.1em, 0.2em)$) node {\ormind{4}} ;
    \draw [magenta] ($(U2.north east) + (0.2em, 0.2em)$) node {\ormind{5}} ;
    \draw [magenta] ($(M2.north east) - (0.1em, 0.2em)$) node {\ormind{6}} ;

    \rules at (-.4, -3.0) {
        \node[rule=1] {A single pipeline transforms the whole dataset at each iteration.};\\
        \node[rule=2] {The output $y_{t,k}$ in the inner loop is a validation measure (e.g. cross-validation).};\\
        \node[rule=3] {The inner loop is initialized with the previous best configuration as prior.};\\
        \node[rule=4] {The inner outputs the best prediction and configuration at iteration $t$.};\\
        \node[rule=5] {$f_M$ returns the best promising configuration w.r.t. the best achievable metric.};\\
        \node[rule=6] {The whole process returns the best configuration to be used in production.};\\
    };

  \end{tikzpicture}
\caption{Two-stage optimization process}
\end{figure}

The proposed two-stage optimization process is illustrated by Figure \ref{fig:architecture}. Due to the sequential nature of the machine learning workflow, it consists in two imbricated loops.
The inner loop performs a cross-validation on $\mathbf X_t$ for a given time budget or until a Cauchy criterion is verified (e.g. the difference in the cross-validation score between two iterations of the inner loop is lower than a threshold $\varepsilon$).

The rationale behind using the previous optimal algorithm configuration $\vec \theta^*_t$ as prior for the next inner loop is that if the ``distance'' between the processed data $\mathbf X_t$ at iteration $t$ and $\mathbf X_{t+1}$ is small, we can expect the new optimal configuration for the algorithm to be rather close, shortening the inner loop computation time if a Cauchy criterion is used. 


Notice that this architecture is independent of the metaoptimizer. Even more, different metaoptimizers can be used for the two stages, and different optimization criteria might be used for each step. For instance, one can imagine optimizing the data pipeline for a fairness criterion and a standard performance metric such as the (cross-validation) accuracy for the algorithm.

\subsection{Policies}

In practical situations, the limiting factor to construct a machine learning workflow is time. Therefore, we are interested in optimization under time constraint, and thus finding how to allocate time between both steps and through iterations. Given a time budget of T, we define different policies of time allocation.

~\\\noindent
{\bf Split policy:} The budget is split between $T_1$ and $T_2$, allocated respectively for the data pipeline configuration search and the algorithm hyperparameter tuning. In the first phase, the metaoptimizer is used during $T_1$ to build the data pipeline. During the second phase, the metaoptimizer is used to configure the algorithm during $T_2$.

Considering the convex combination $T = (1 - \omega) T_1 + \omega T_2$, for $\omega = 0$, no hyperparameter tuning is done because the whole budget is spent on building the data pipeline. Conversely, for $\omega =1.0$, the process is fully dedicated to tune hyperparameters of the algorithm.\\

\noindent
{\bf Iterative policy:} Each step alternates during a short runtime of $t$ seconds. The best configuration found during a step is reused during the next step, iteratively until the total budget is expired. It is relatively fast for the metaoptimizer to find a better pipeline configuration than the baseline, but then, stagnates. Therefore, the time would be better allocated to the search for a better algorithm configuration. If the data pipeline was modified, the data used to train the algorithm changes. Those variations might help during the hyperparameter tuning to explore a new region of the search space compared to the previous iteration.\\

\noindent
{\bf Adaptive policy:} This policy reuses the iterative policy. However, the time allocation is not fixed per iteration. If during an iteration the cross-validation score improved, the time allocated to the next iteration of the same type (data pipeline or algorithm configuration) is multiplied by two. Conversely, if after two iterations of the same type, the score is not improved, the allocated time for this type of iteration is divided by two.\\

\noindent
{\bf Joint policy:} This policy simply uses the union of both search space. In other words, it is equivalent to what is usually done in practice by current meta-optimizers, i.e., searching over the whole space of machine learning pipelines. \\

\section{Experimental Setting}
\label{sec:exp_1}


\subsection{Experimental setting}
\label{sec:protocol_1}

\noindent
{\bf Datasets and algorithms. ~~} We performed the experiments on 3 datasets: Wine\footnote{\url{https://scikit-learn.org/stable/modules/generated/sklearn.datasets.load_wine.html}}, Iris\footnote{\url{https://scikit-learn.org/stable/modules/generated/sklearn.datasets.load_iris.html}} and Breast\footnote{\url{https://scikit-learn.org/stable/modules/generated/sklearn.datasets.load_breast_cancer.html}}\footnote{The choice of small datasets is justified by the need to know the optimal score in the search space to effectively evaluate the metaoptimizer and policies results.}. We used 4 classification algorithms: SVM, Random Forest, Neural Network and Decision Tree. The implementation is provided by Scikit-Learn \cite{scikit-learn}.

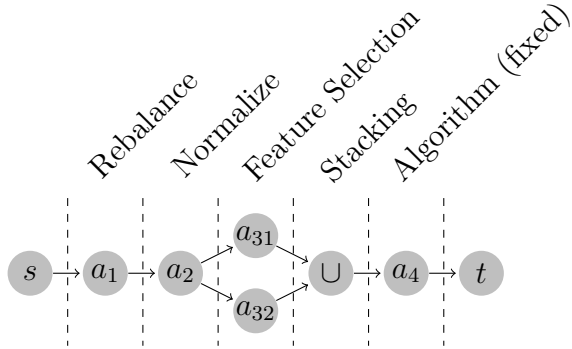
\begin{figure}[h!]
\center
\begin{tikzpicture}[shorten >=1pt,->]
  \tikzstyle{vertex}=[circle,fill=black!25,minimum size=17pt,inner sep=0pt]

  \node[vertex] (S) at (0,0) {$s$};
  \node[vertex] (O1) at (1,0) {$a_1$};
  \node[vertex] (O2) at (2,0) {$a_2$};
  \node[vertex] (O31) at (3,0.5) {$a_{31}$};
  \node[vertex] (O32) at (3,-0.5) {$a_{32}$};
  \node[vertex] (U) at (4,0) {$\cup$};
  \node[vertex] (A) at (5,0) {$a_4$};
  \node[vertex] (T) at (6,0) {$t$};

  \draw (S) -- (O1);
  \draw (O1) -- (O2);
  \draw (O2) -- (O31);
  \draw (O2) -- (O32);
  \draw (O31) -- (U);
  \draw (O32) -- (U);
  \draw (U) -- (A);
  \draw (A) -- (T);

  \draw [-,dashed] (0.5, 1) -- (0.5, -1);
  \draw [-,dashed] (1.5, 1) -- (1.5, -1);
  \draw [-,dashed] (2.5, 1) -- (2.5, -1);
  \draw [-,dashed] (3.5, 1) -- (3.5, -1);
  \draw [-,dashed] (4.5, 1) -- (4.5, -1);
  \draw [-,dashed] (5.5, 1) -- (5.5, -1);

  \node[rotate=45] at (1.5,2) {Rebalance};
  \node[rotate=45] at (2.6,2) {Normalize};
  \node[rotate=45] at (4,2.4) {Feature Selection};
  \node[rotate=45] at (4.5,1.8) {Stacking};
  \node[rotate=45] at (6,2.4) {Algorithm (fixed)};

\end{tikzpicture}
\caption{Pipeline topology created for the experiments.}
\label{fig:topology}
\end{figure}

~\\\noindent
{\bf Data pipeline search space. ~} We created a {\it pipeline prototype} made of three layers: ``rebalance'' (handling imbalanced dataset), ``normalize'' (scaling features), ``features'' (feature selection or dimension reduction). For each step, we selected few possible concrete operators with a specific configuration space summarized in Table~\ref{table:pipeline_search_space}. The topology of the pipeline is defined by Figure \ref{fig:topology}. Each node can be instantiated with an operator or left empty. When both ``rebalance'' slots are instantiated, the final vector is obtained by stacking the output of both operators.  
There is a total of 4750 possible pipeline configurations. This is roughly the same as in \AUTOSKLEARN~or \MOSAIC, and about 5 times less than \AUTOWEKA. 

\setlength{\fboxrule}{0.8pt}

~\\\noindent
{\bf Algorithm hyperparameter search space. ~~} For each algorithm, we defined a reasonable hyperparameter space with approximately the same size as the data pipeline space. The search space size contains 4800 elements for Random Forest and Decision Tree, 1944 for Neural Network and 768 for SVM. 

~\\\noindent
{\bf Metaoptimizer. ~~} We selected \HYPEROPT~\cite{bergstra2015hyperopt} as metaoptimizer as it has been shown to perform better than alternatives for high dimensional problems \cite{eggensperger2013}. A 10-fold cross-validation is used to assess the pipeline performances.

~\\\noindent
{\bf Resources. ~~}The machine used for the experiments is equipped with an Intel i7-6820HQ and 32GB RAM.

\subsection{Protocol and goals}

We decompose the experiments in two parts. First, given a restricted budget, we want to assess the respective influence of searching for a data pipeline and configuring the algorithm. The second part focuses on the optimization with a time budget and the evaluation of two-stage optimization process, and in particular the different policies. All experiments are reproducible and the source-code is documented and available in a dedicated GitHub repository\footnote{\url{https://github.com/aquemy/DPSO\_experiments}}.

~\\\noindent
{\bf Experiment 1. } We would like to quantify the impact of each phase on the final result. For this, we proceed in two steps. First, we perform an exhaustive search in the data pipeline search space, followed by a search using \HYPEROPT~with a budget of 100 configurations to explore (about 2\% of the configuration space). The algorithm uses the default configuration of its implementation. In a second step, we perform the same for hyperparameter tuning with the baseline data pipeline. Those two steps have been repeated for each algorithm and each dataset. We report the density of configurations depending on the accuracy, both for the exhaustive search and for the restricted budget. 

~\\\noindent
{\bf  Experiment 2. } We ran the two-stage optimization process for $T=300$ seconds, for each dataset, each method and each policy. For the {\bf split policy}, we performed the experiment for each $\omega \in \{0, 0.1, ..., 0.9, 1 \}$ to show the effect of different allocations between the two stages. For the {\bf iterative policy}, we setup the iteration runtime to 15 seconds. Similarly, the default iteration time for the {\bf adaptive policy} has been fixed to 15 seconds.

\section{Results}
\label{sec:results_1}

\subsection{Experiment 1}

\begin{figure}[h!]
\includegraphics[scale=0.35]{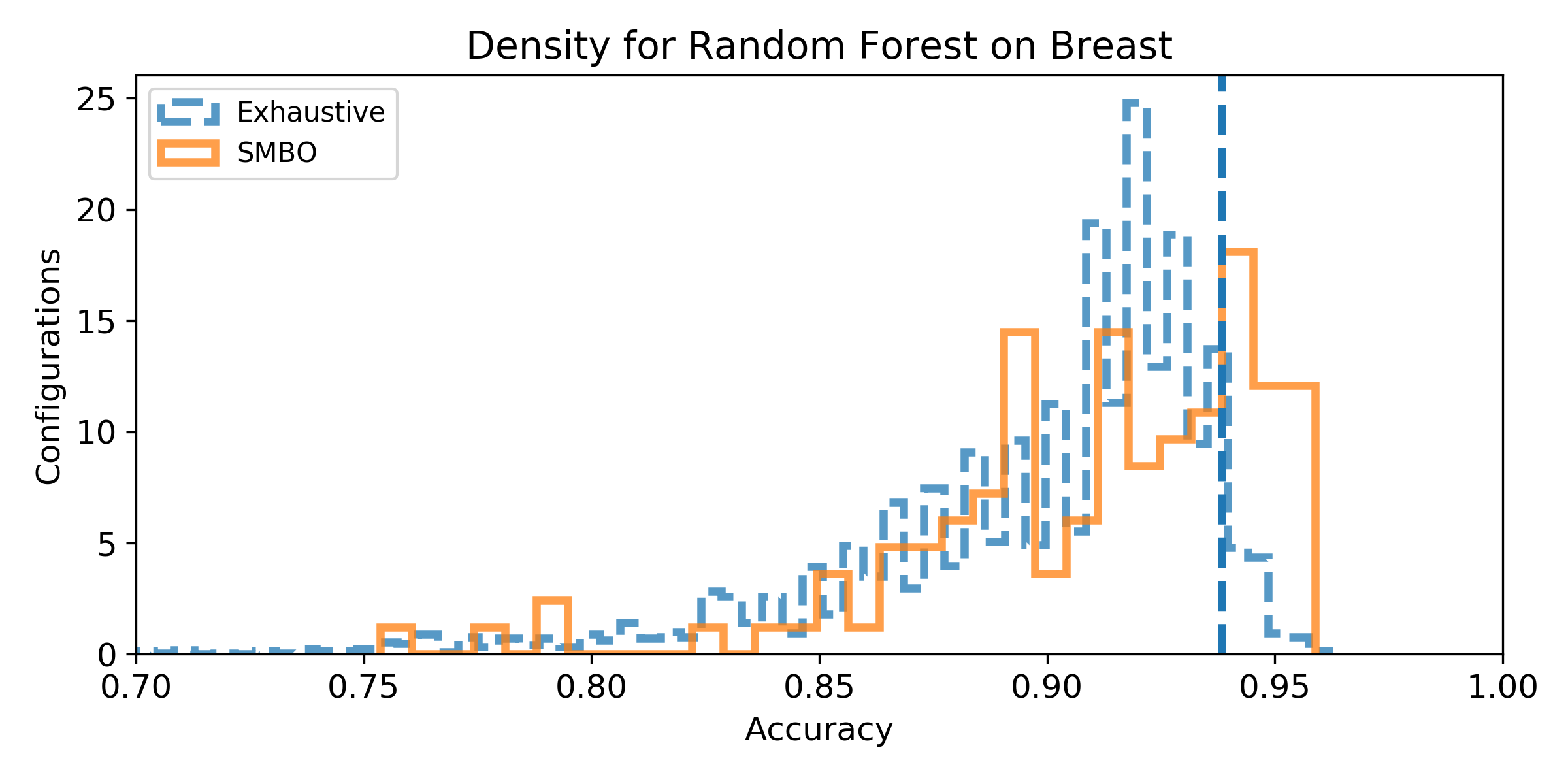}
\includegraphics[scale=0.35]{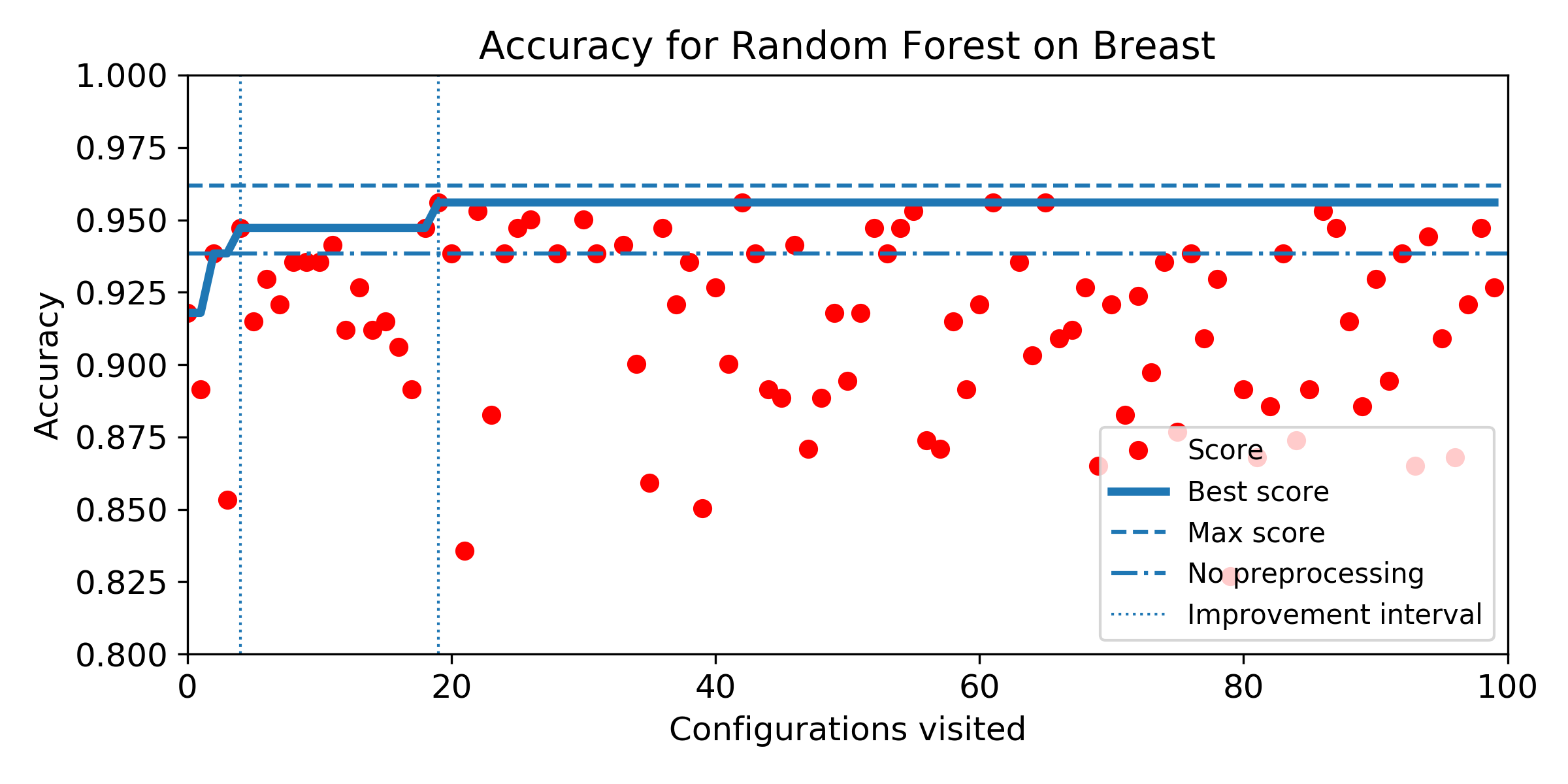}\\
\includegraphics[scale=0.35]{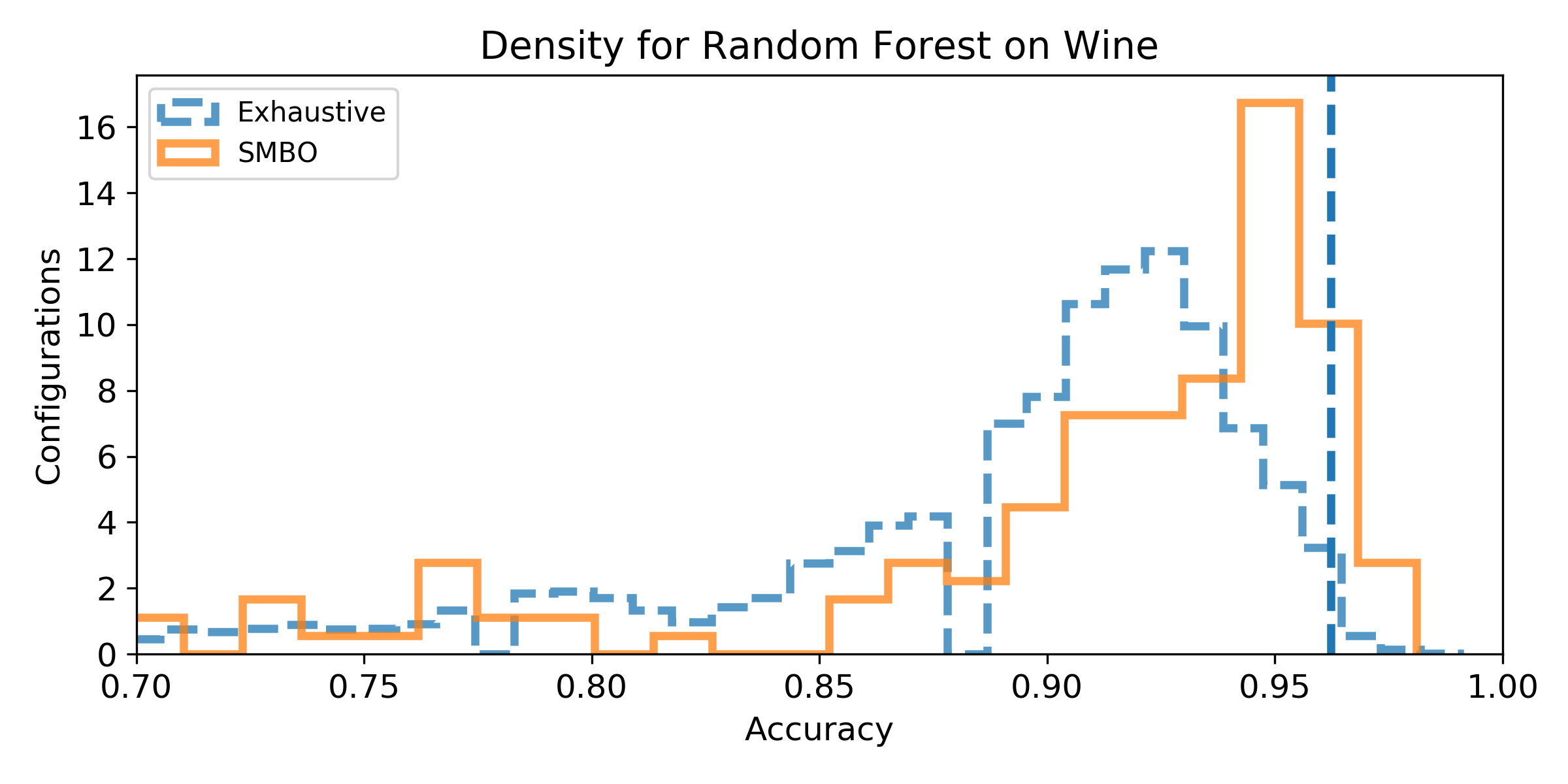}
\includegraphics[scale=0.35]{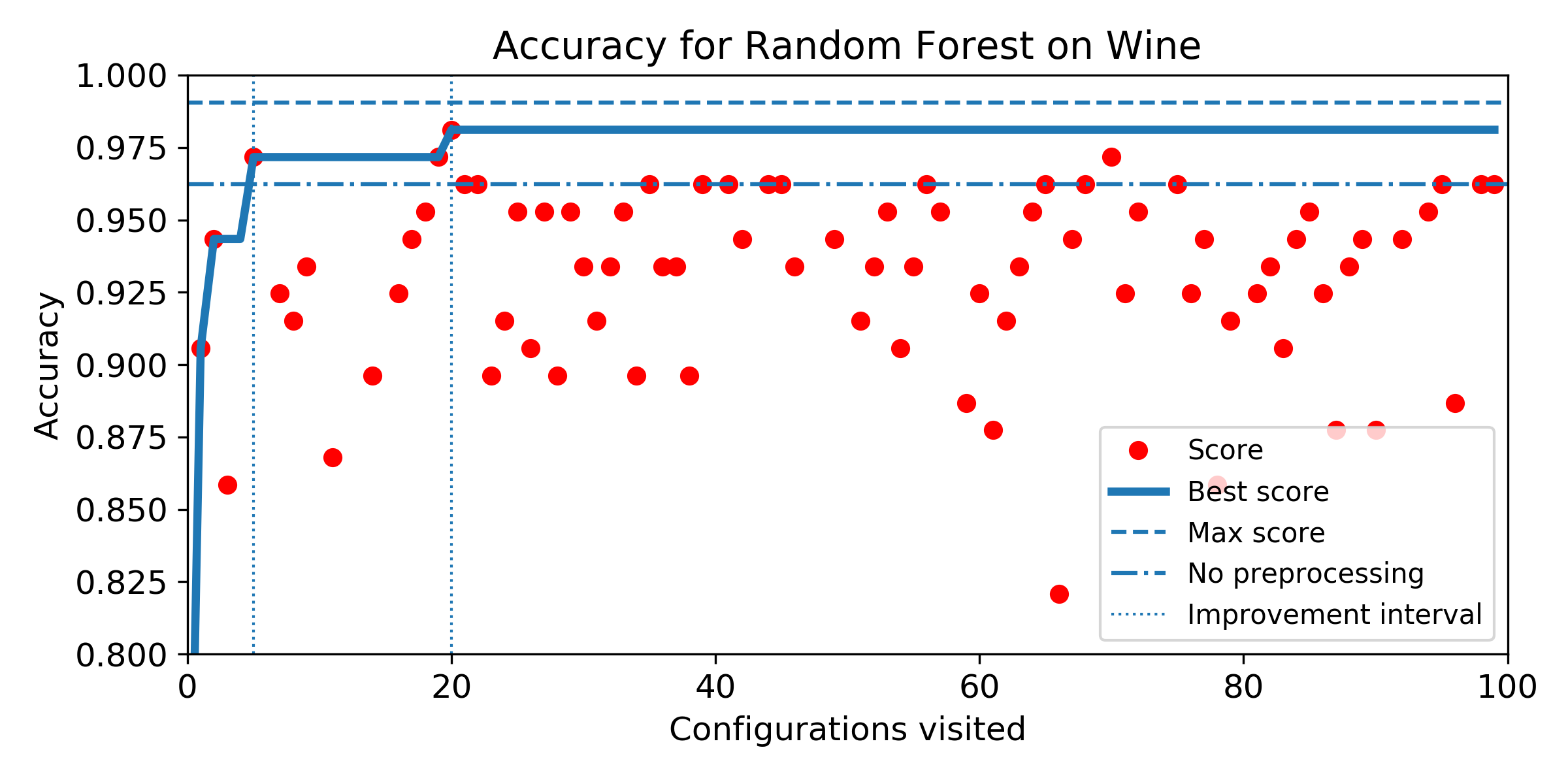}
\caption{Density of pipeline configurations (left). The vertical line represents the baseline score. Evolution of the accuracy iteration after iteration (right).}
\label{example1}
\end{figure}

Figure \ref{example1} provides the result obtained with Random Forest on Breast and Wine. A summary of the results is provided by Table \ref{table:exp1_results}. All results are qualitatively similar.
Figure \ref{example1}, on Breast (top part), shows that the baseline score is 0.9384 and the best score 0.9619 i.e. an error reduction of 38\% is achievable in the pipeline search space. Similarly, on Wine (bottom part), the best accuracy is 0.9906, i.e. a reduction of 25\% of the error rate. Most configurations deteriorate the baseline score. However, \HYPEROPT~is skewed towards better configuration compared to the exhaustive search. It indicates \HYPEROPT~has a better probability to find a good configuration than random search. The right parts show that \HYPEROPT~starts to improve the baseline score after only 4 iterations and reached its best configuration after 19 iterations on Breast (resp. 5 and 20 for Wine).

Table \ref{table:exp1_results} shows that similar results are obtained for all methods on all datasets. \HYPEROPT~always found a better pipeline than the baseline, in at most 17 iterations. In average, the best score is achieved around 20 iterations (excluding Decision Tree on Iris and Breast). Decision Tree was able to reach the optimal configuration on Iris (resp. Wine) after 1 (resp. 5) iterations. In general, the score in the normalized score space belongs to $[0.9780, 1.000]$. To summarize, in average, with 20 iterations (0.42\% of the search space) \HYPEROPT~is able to decrease the error by 58.16\% compared to the baseline score and found configurations that reach 98.92\% in the normalized score space.

\begin{table}
\caption{Optimization results.}
\label{table:exp1_results}

\centering
\resizebox{1\textwidth}{!}{%

\begin{tabular}{l|c|cc|cccc|cc}
\toprule
  \multicolumn{2}{c}{} & \multicolumn{2}{|c}{Exhaustive} & \multicolumn{4}{|c}{Cross-validation Score} & \multicolumn{2}{|c}{Imp. Interval}\\
\midrule
& Baseline & DP & Algo. & DP & DP (norm.) & Algo. & Algo. (norm.) & DP & Algo.\\
\midrule
\multicolumn{9}{c}{Iris}\\
\midrule
SVM & 0.9667 & \bf 0.9889 & - & 0.9778 & 0.9831 & \bf 0.9866 & - &  $[11,11]$ & $[2,10]$ \\
Random Forest & 0.9222 & \bf 0.9778 & 0.9667  &\bf  0.9667 & 0.9828 & \bf 0.9667 & 1.0000 &  $[8,27]$ & $[1,70]$ \\
Neural Net & 0.9667 & \bf 0.9889 & 0.9778 & \bf 0.9778 & 0.9831 & 0.9667 & 0.9820 &  $[17,17]$ & $[1,11]$ \\
Decision Tree & 0.9222 & \bf 0.9889 & 0.9667  & \bf 0.9889 & 1.0000 & 0.9667 & 1.0000 &  $[1,83]$ & $[12,36]$\\
\midrule
\multicolumn{9}{c}{Breast}\\
\midrule
SVM & 0.9501 & \bf 0.9765 & - & \bf 0.9765 & 1.0000 & 0.9474 & - &  $[12,20]$ & - \\
Random Forest & 0.9384 & 0.9619 & \bf 0.9765 & 0.9560 & 0.9780 & \bf 0.9685 & 0.8982 &  $[4,19]$ & $[4,46]$\\
Neural Net &  0.9326 & \bf 0.9765 & 0.9472 & \bf 0.9707 & 0.9903 &  0.9175 & 0.9600 &  $[1,7]$ & $[3,13]$ \\
Decision Tree & 0.9296 & 0.9619 & \bf 0.9648 & \bf 0.9589 & 0.9900 & 0.9527 & 0.9605 &  $[0,67]$ & $[13,23]$\\
\midrule
\multicolumn{9}{c}{Wine}\\
\midrule
SVM & 0.9151 & \bf 1.0000 & 0.9728 & \bf 0.9906 & 0.9811 & 0.9728 & 1.0000 &  $[3,13]$ & $[1,3]$ \\
Random Forest & 0.9623 & \bf 0.9906 & \bf 0.9906 & 0.9811 & 0.9818 & \bf 0.9906 & 1.0000 &  $[5,20]$ & $[1,23]$\\
Neural Net & 0.9057 & \bf 0.9906 & 0.9434 & \bf 0.9906 & 1.0000 & 0.9245 & 0.9778 &  $[1,25]$ & $[1,13]$  \\
Decision Tree & 0.9057 & \bf 0.9811 & 0.9528 & \bf 0.9811 & 1.0000 & 0.9339 & 0.9671 &  $[5,35]$ & $[11,75]$\\
\bottomrule
\end{tabular}
} 

\begin{flushleft} \small
DP and Algo. represents respectively the Data Pipeline phase and the algorithm phase. The column norm. is the cross-validation score normalized within the search space. The last column is the interval where the left bound is the number of configurations required for \HYPEROPT~to improve the baseline score, and the right, the number of configurations before reaching the best score.
\end{flushleft}
\end{table}

Exhaustive results for the algorithm step are better for Random Forest and Decision Tree on Breast only and similar to the data pipeline one only for Random Forest on Wine. In all other nine cases, the optimal score reachable in the search space is better for the data pipeline phase. 

Regarding the score after 100 iterations, \HYPEROPT~found a better accuracy for the data pipeline construction in all cases, except for Random Forest on all datasets, and SVM on Iris dataset.

In conclusion, for a similarly large and realistic search space, and a given number of configurations to explore, a good data pipeline with a default algorithm configuration provides better results than a tuned algorithm with no pre-processing operations. A notable exception is Random Forest that largely benefits from hyperparameter tuning.

\subsection{Experiment 2}

\begin{figure}[h!]
\center
\includegraphics[scale=0.35]{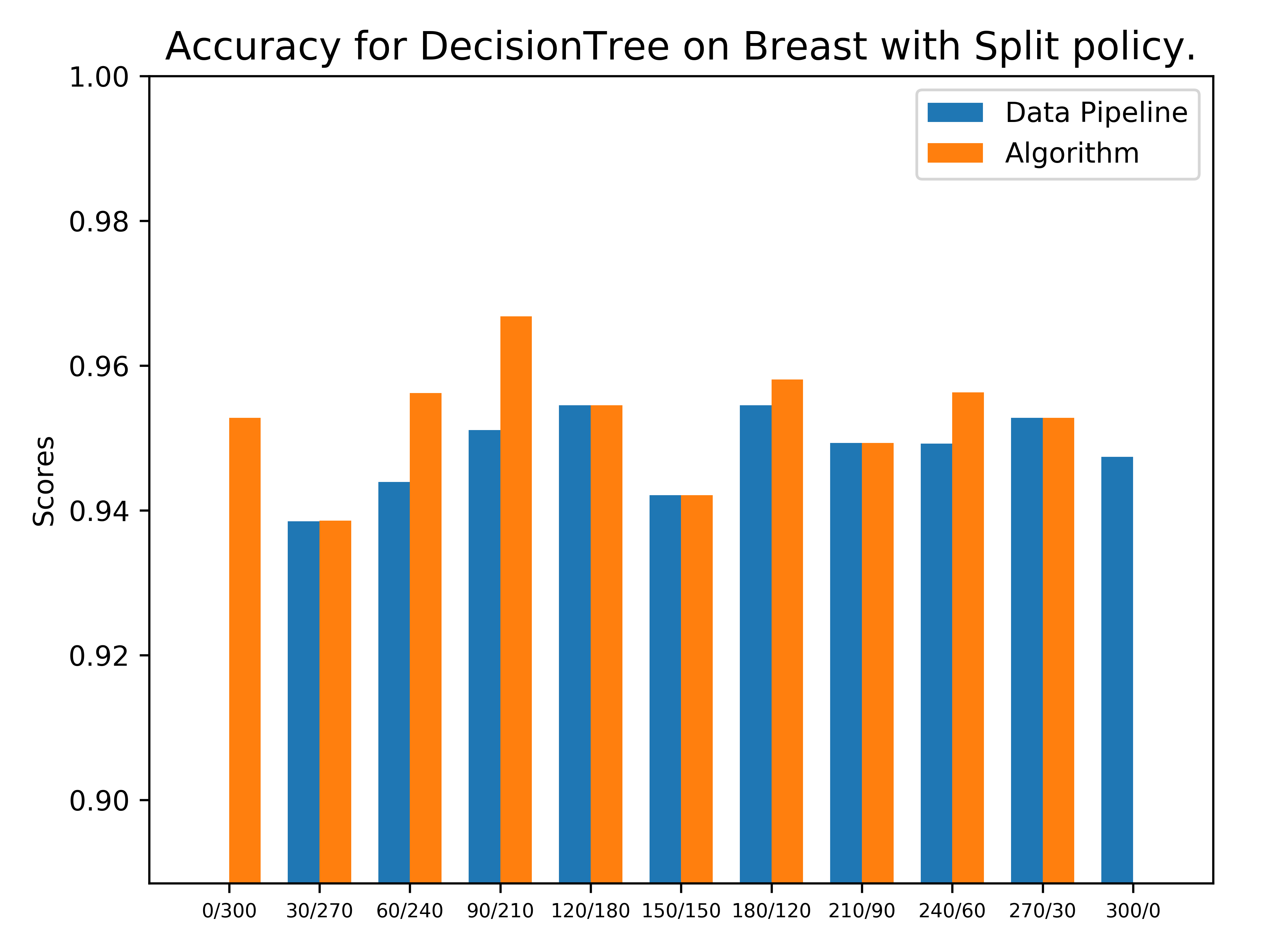}
\includegraphics[scale=0.35]{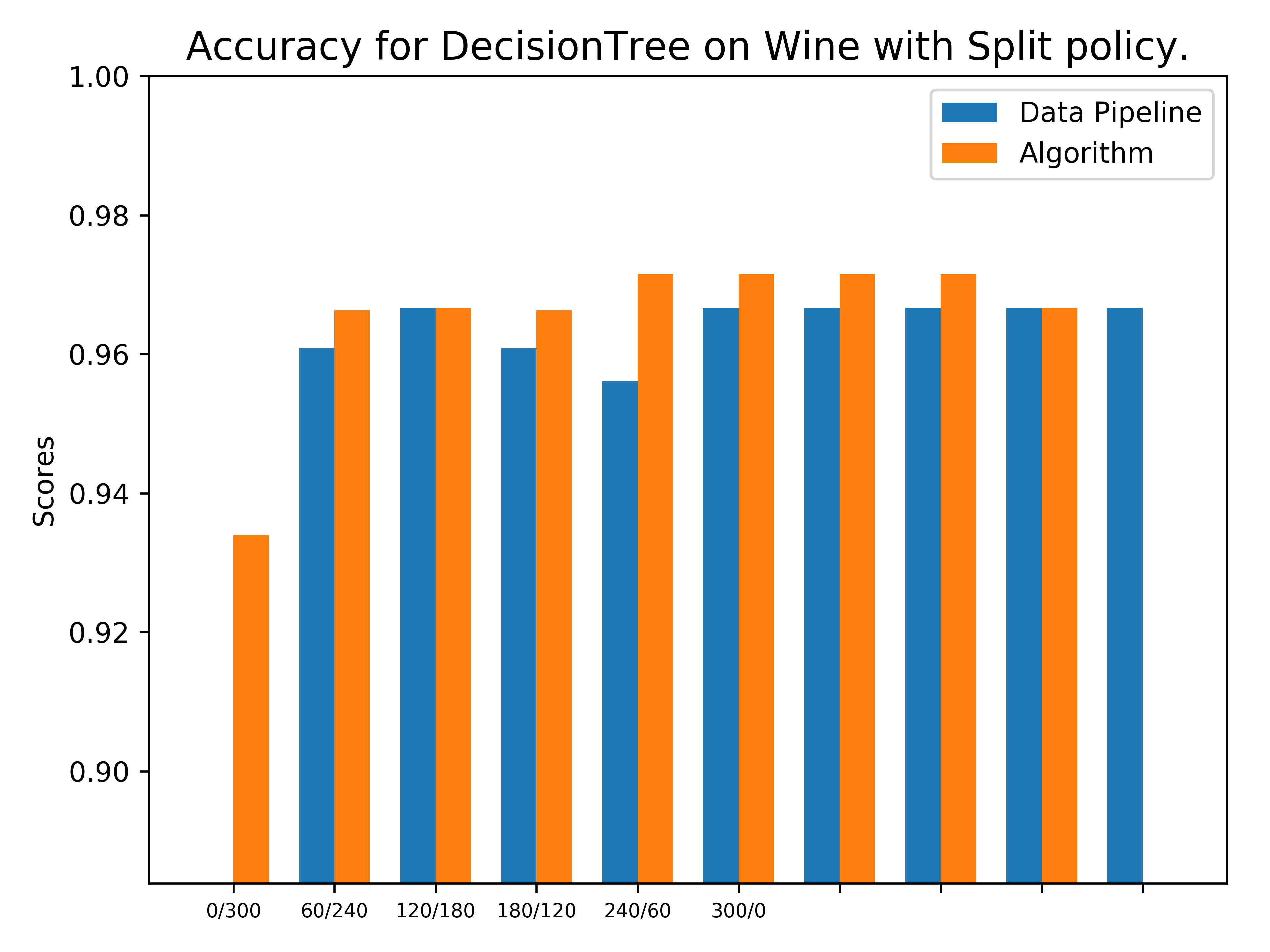}
\includegraphics[scale=0.35]{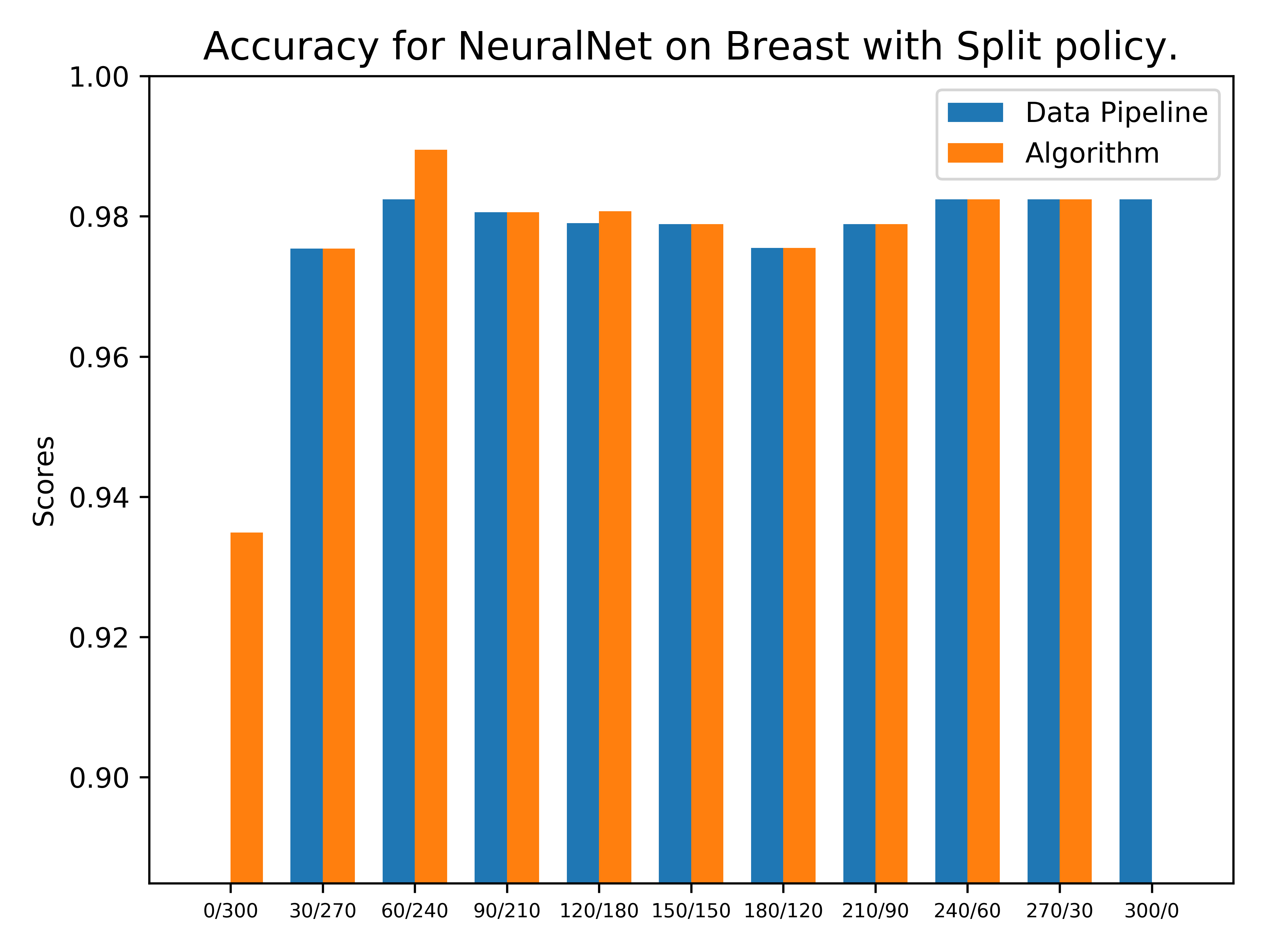}
\includegraphics[scale=0.35]{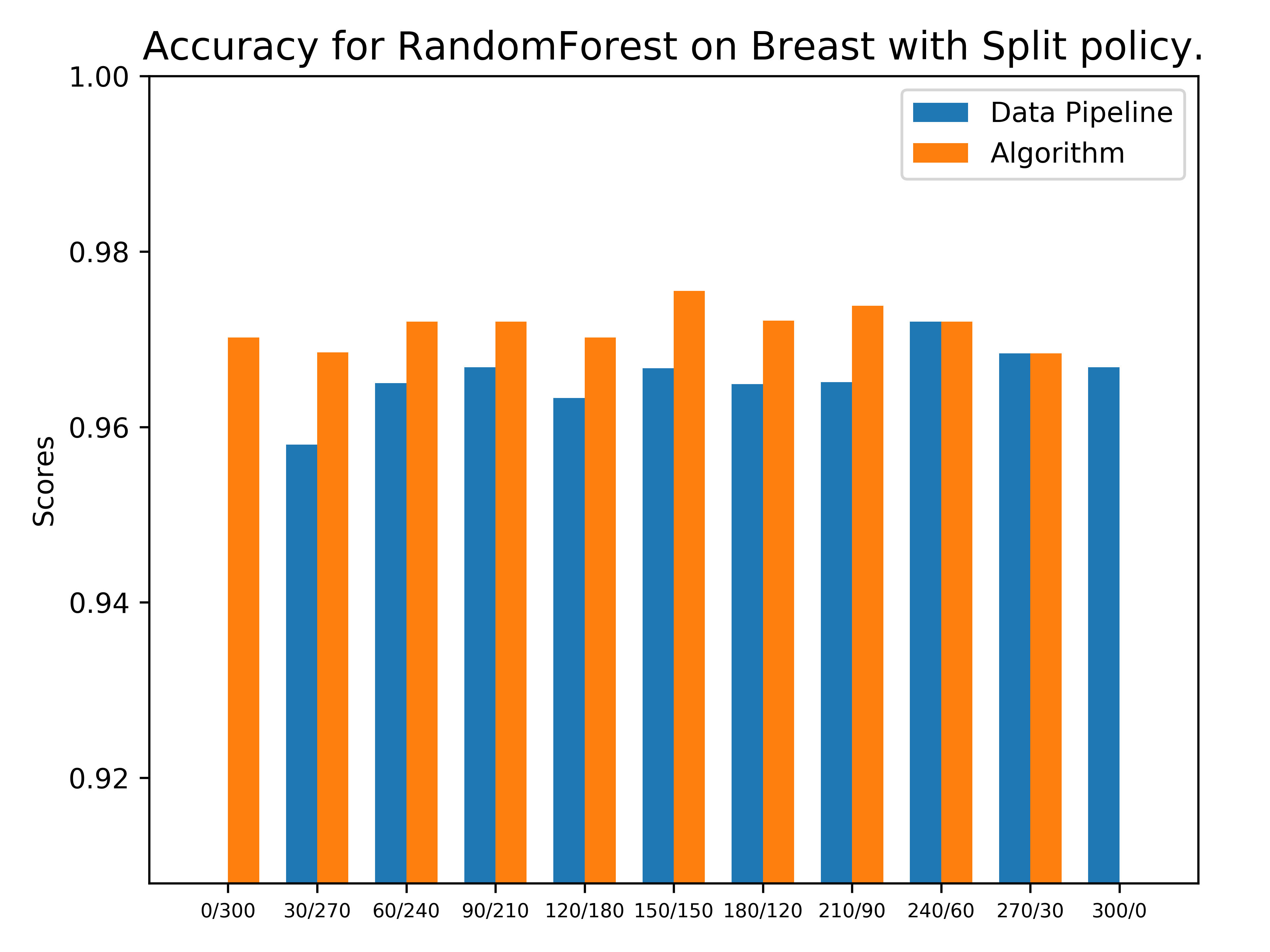}
\caption{Accuracy depending on the time spent on each phase of the optimization process.}
\label{policy_split}
\end{figure}

Figure \ref{policy_split} shows the accuracy depending on different time allocations between each phase. By construction, the accuracy reached on the second phase can be only equal or higher to the best accuracy reached by the first phase. For the same method, depending on the dataset it might be better to allocate the whole budget on a phase or another. This is the case for Decision Tree: allocating the whole budget to the algorithm configuration on Breast returns better results than spending the whole budget on the data pipeline construction. The exact opposite is observed on Wine.
In general, the best accuracy is obtained by a tradeoff between the two phases which seems to depend both on the algorithm and the dataset. Therefore, more advanced time allocation policies such as iterative or adaptive may help. Learning to predict optimal time allocation for a new dataset and algorithm based on previous runs is left for future work.

\begin{figure}[]
\includegraphics[scale=0.35]{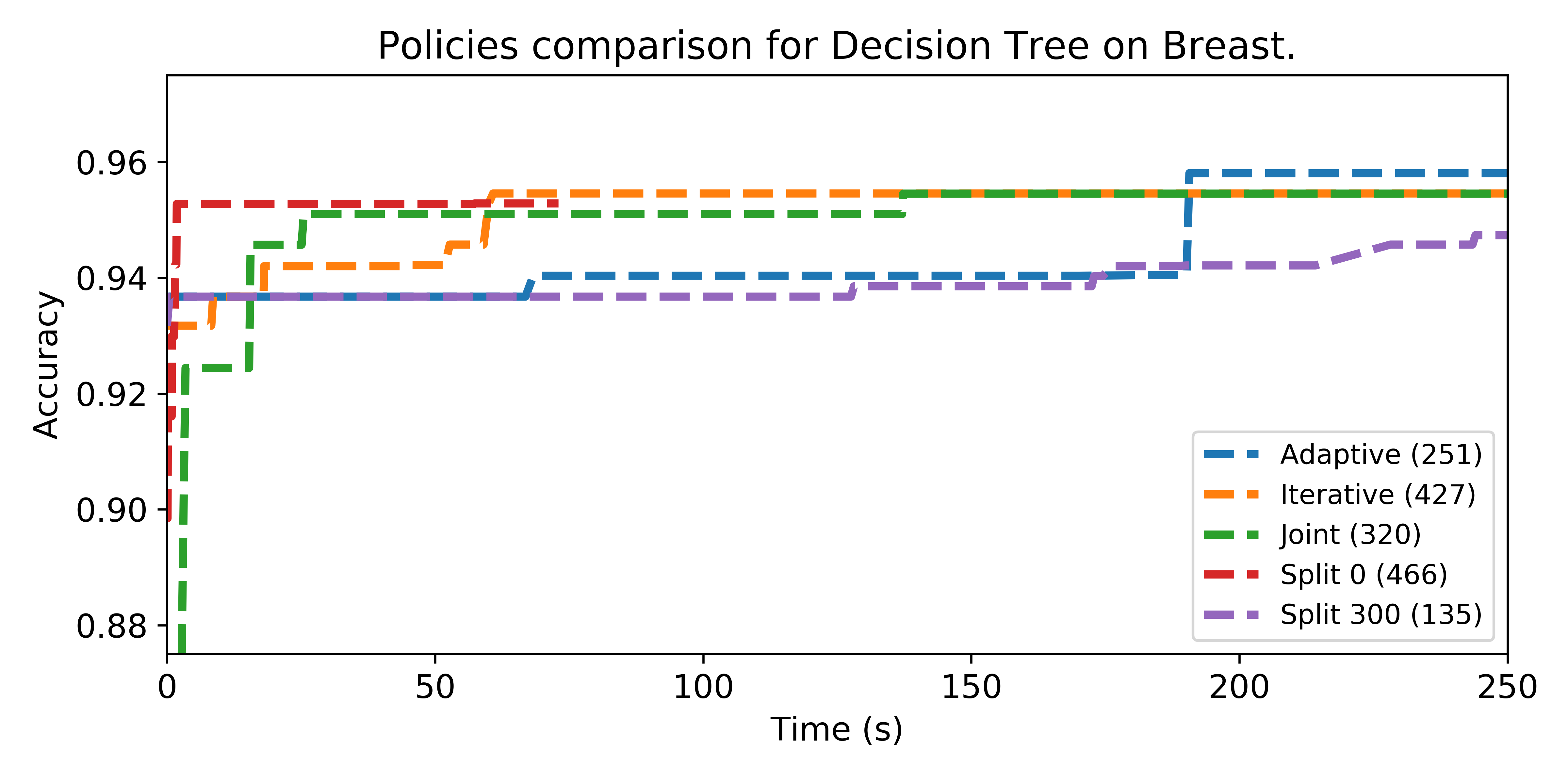}
\includegraphics[scale=0.35]{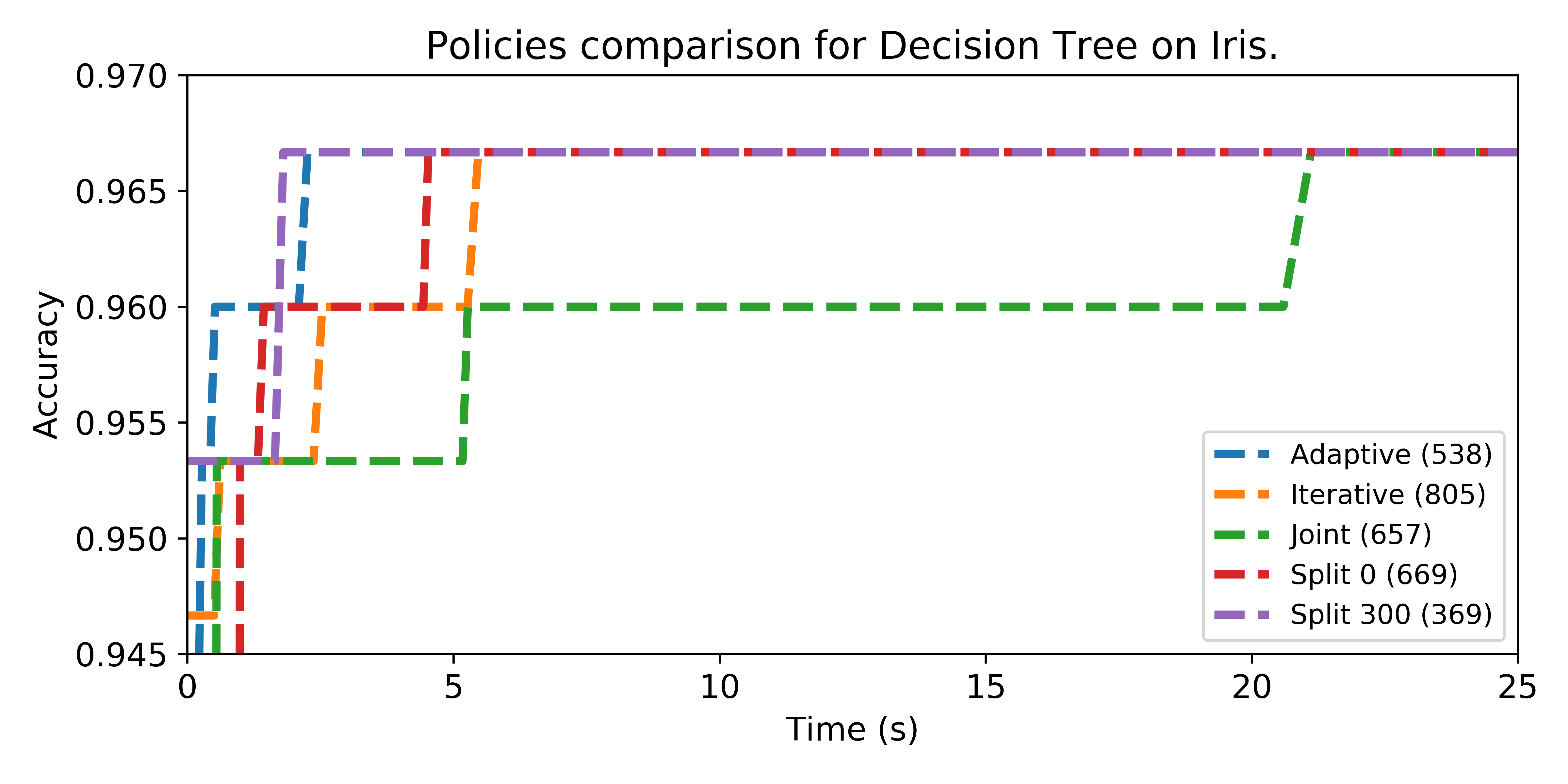}
\includegraphics[scale=0.35]{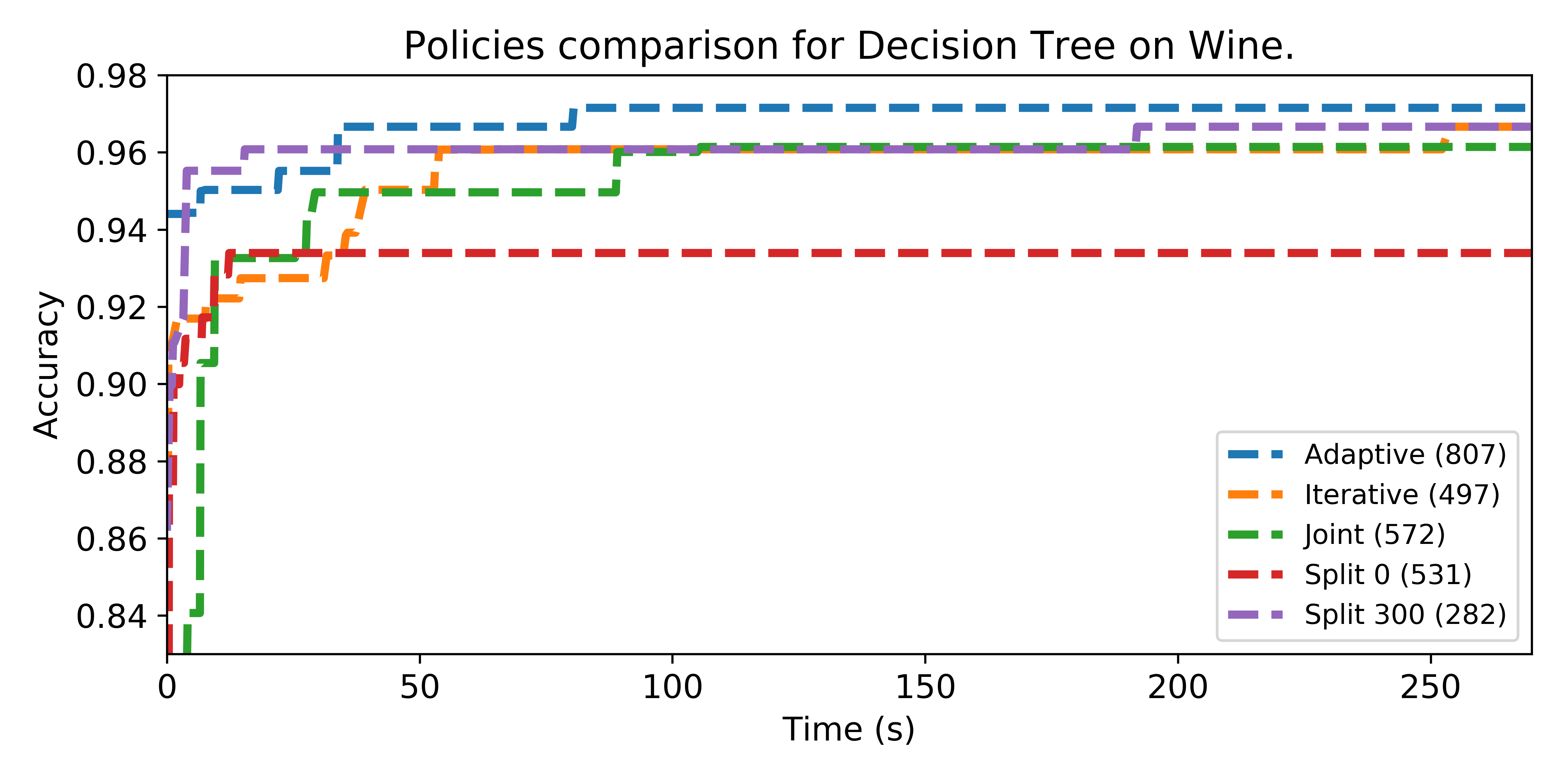}
\includegraphics[scale=0.35]{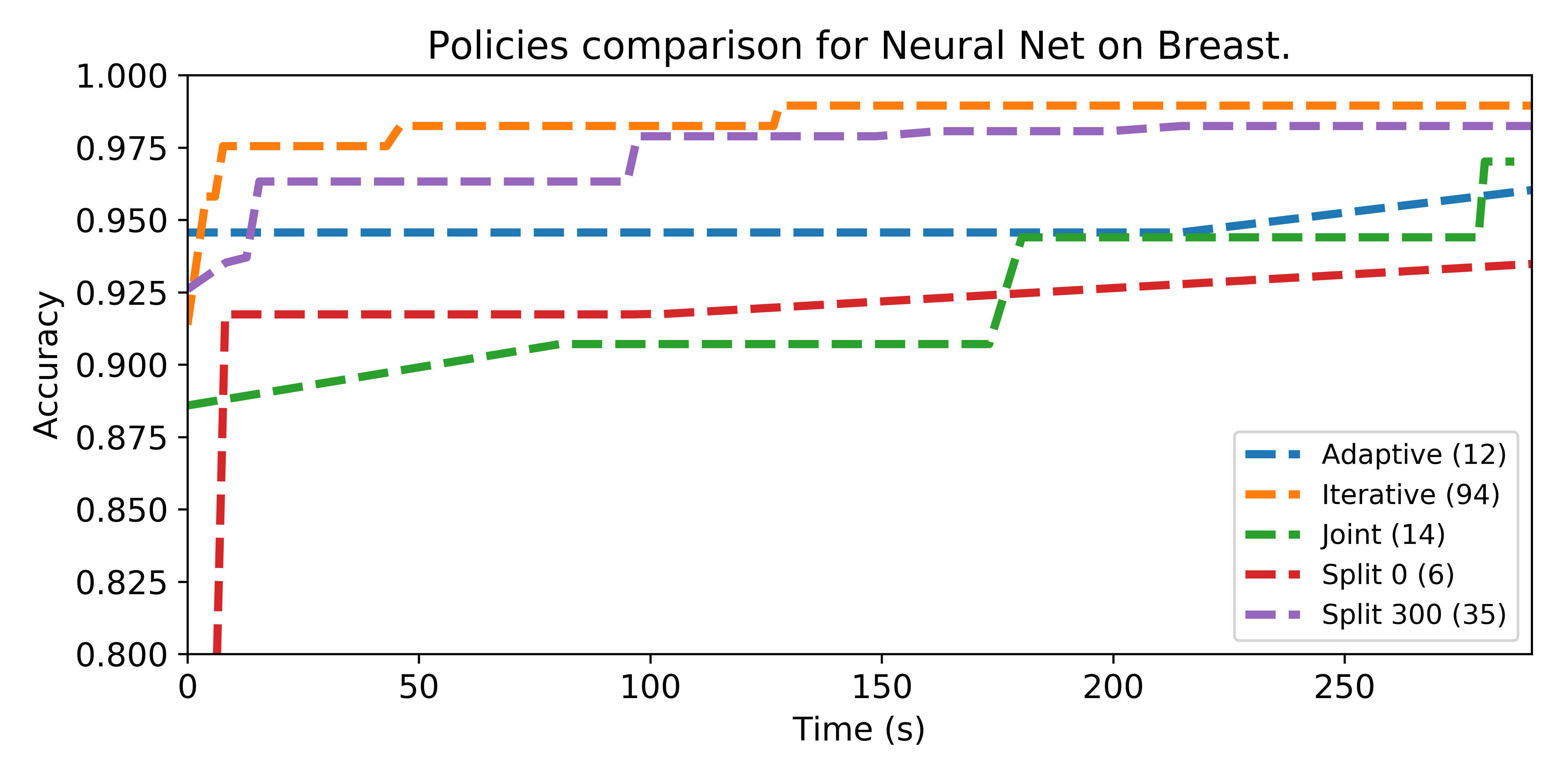}
\includegraphics[scale=0.35]{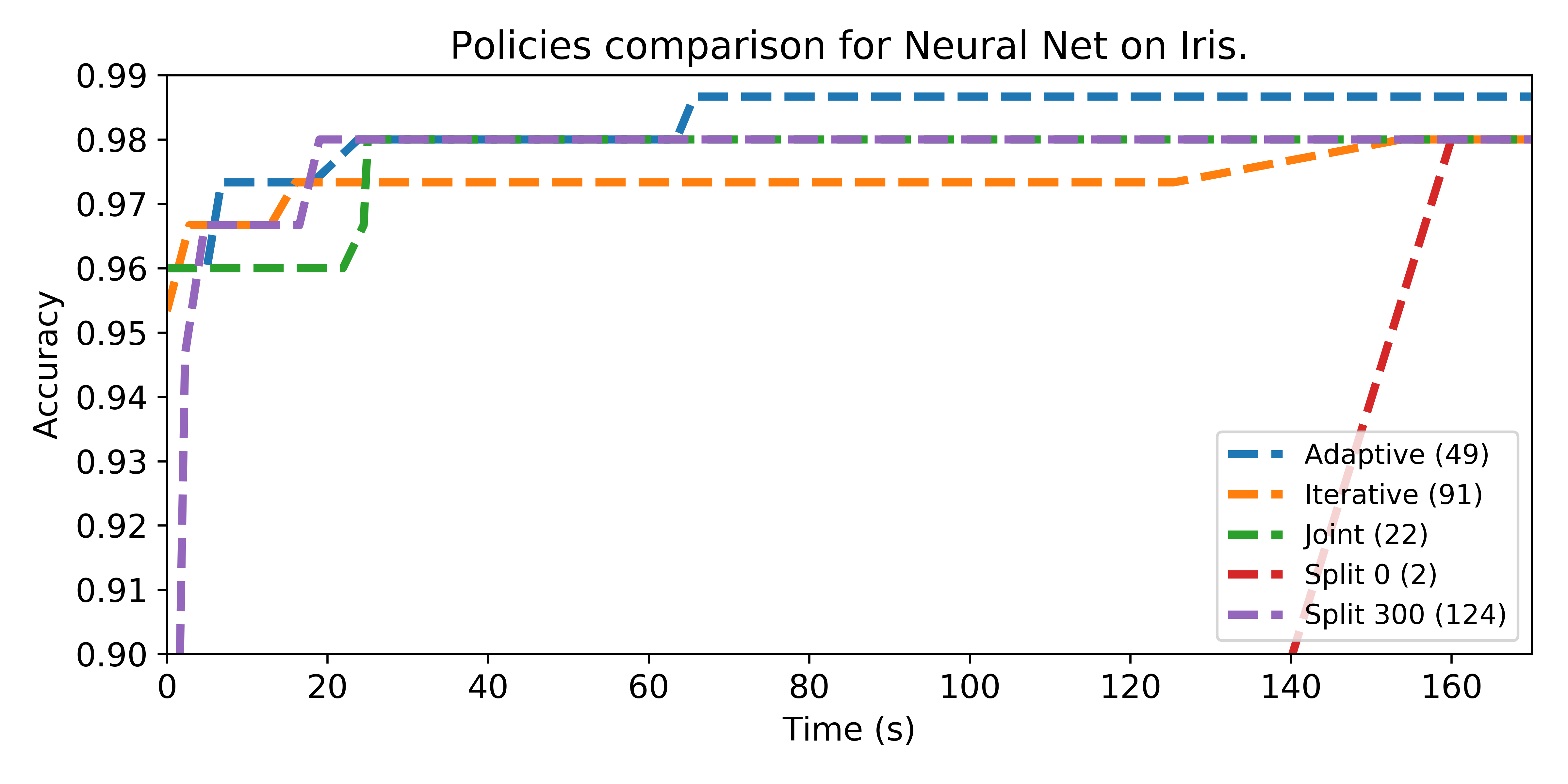}
\includegraphics[scale=0.35]{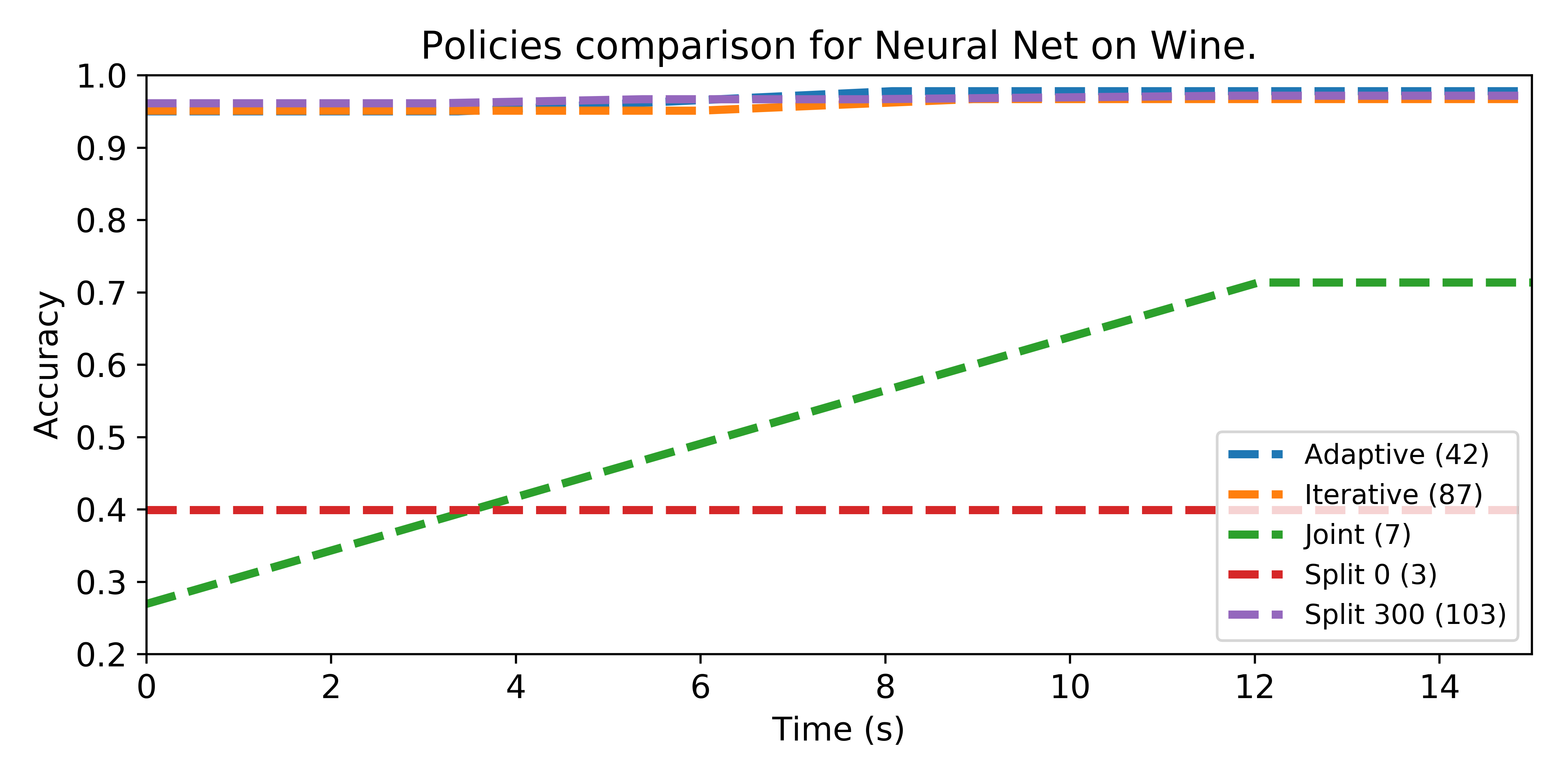}
\includegraphics[scale=0.35]{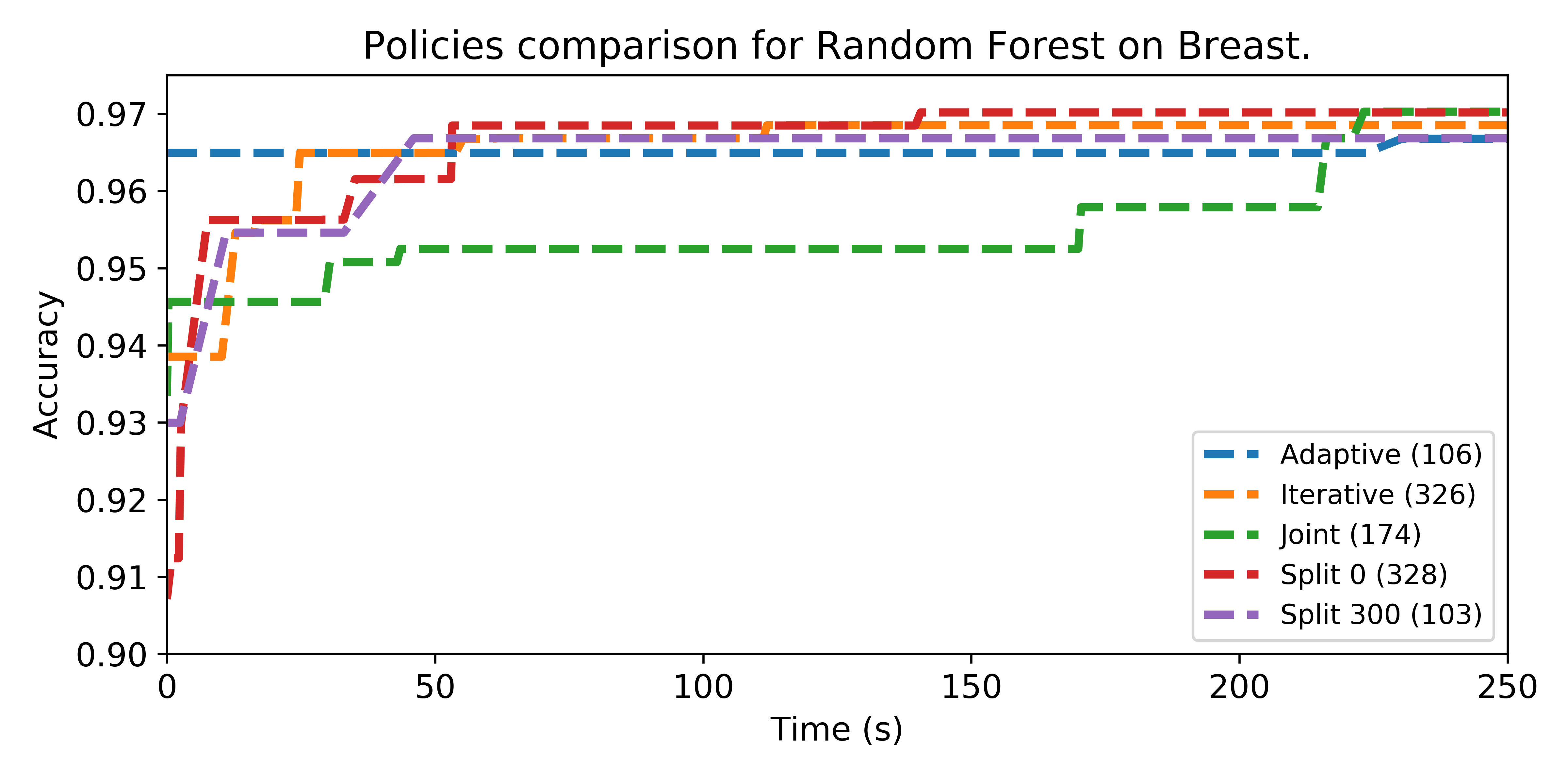}
\includegraphics[scale=0.35]{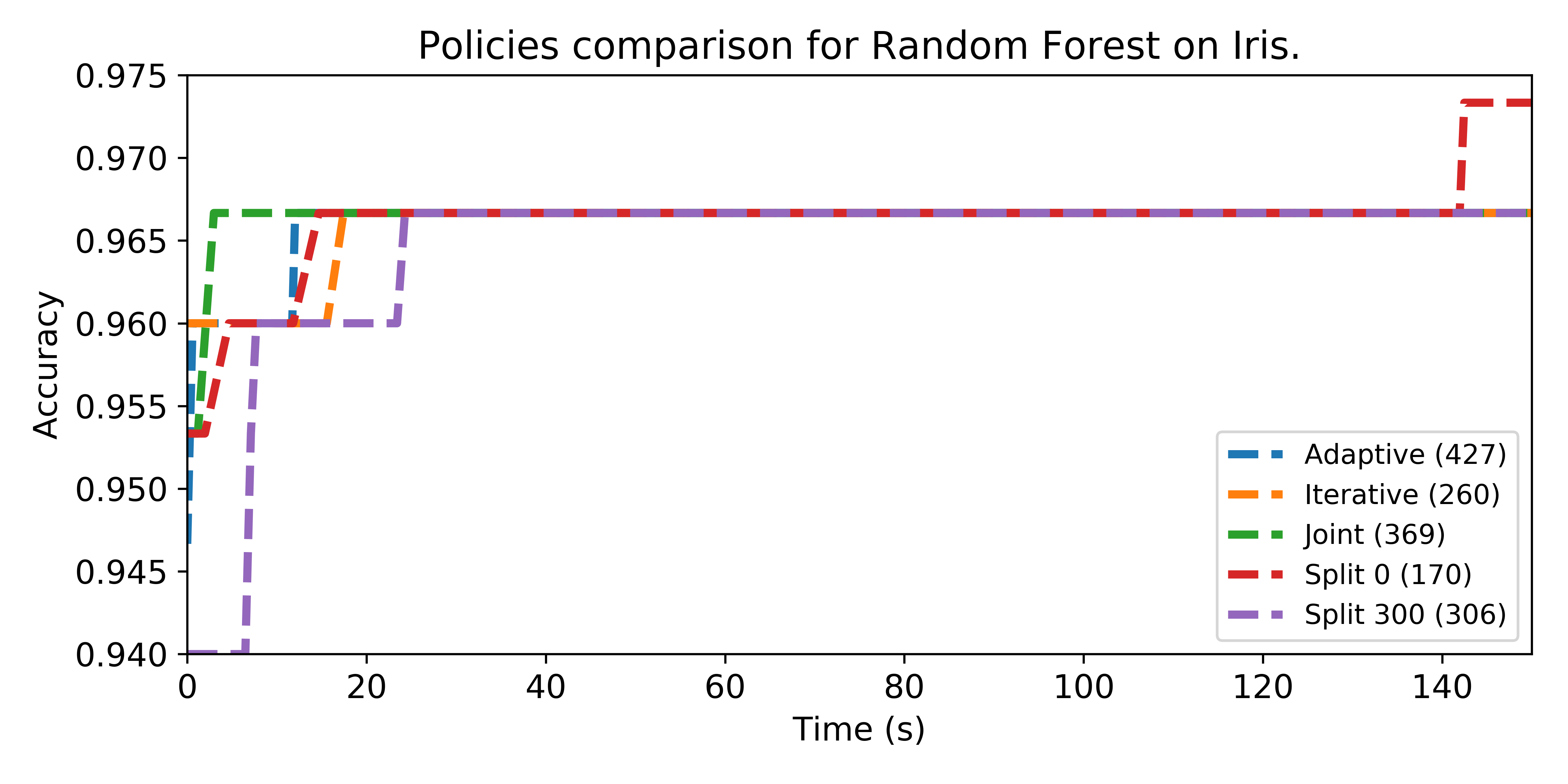}
\includegraphics[scale=0.35]{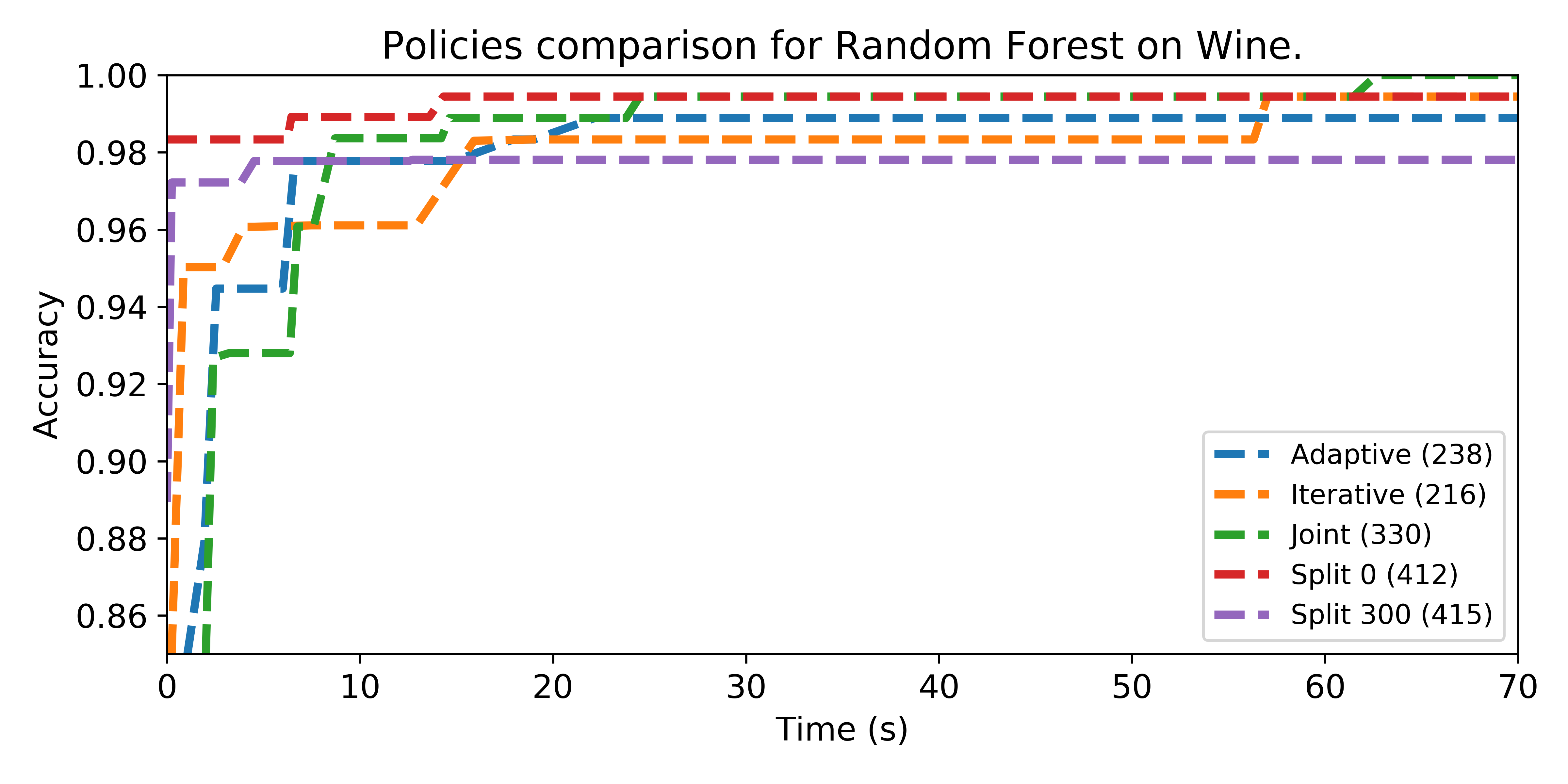}
\caption{Evolution of the best score in time for different policies. Split 300 and Split 0 are respectively fully dedicated to the data pipeline selection or the algorithm configuration. In brackets is indicated the number of configuration visited.}
\label{policy_comparison}
\end{figure}

We reported the evolution of the best accuracy through time in Figure \ref{policy_comparison}.
Joint policy returns a lower or equal accuracy in all cases but Random Forest on Wine. In general, it is far slower to reach a good score compared to other policies except for Random Forest on Iris where it is the first policy to reach the plateau. It is particularly visible for Decision Tree on Iris where all policies behave the same and reach the same score. In this case, Joint policy took about 20 seconds to reach the best score against 5 seconds for the second worst and 2 seconds for the best.

The results of Split 0 policy with Neural Net are really poor for all datasets. The reason is the little number of configurations visited due to the time required to train the algorithm without feature selection. On the contrary, Split 0 performs relatively well for Random Forest on all datasets. This seems to indicate that the importance of both optimization stages are different depending on the algorithm.

The Adaptive policy reaches the best score in five cases, and Iterative policy in three cases. In general, at the exception of Decision Tree on Wine, the Adaptive policy manages to reach similar or better scores with less configurations sampled. The best examples are Decision Tree on Breast where the Adaptive policy reaches the best score but samples only 251 configurations against 320 and 427 for the second best policies, as well as Neural Network on Iris with 49 configurations versus 22 and 124 for a lower score.

From those results, it appears that the Iterative and Adaptive policies are the most robust ones to quickly provide good results for any method and any dataset. However, depending on the specificities of the method and the dataset, spending 100\% of the time on one phase or the other might be better. In this case and in theory, the adaptive policy is equivalent to one of those Split policy after some time, i.e. the optimization process spends most of its time on one phase.

\section{Algorithm-specific configuration}
\label{sec:exp_2}

A pipeline is obviously specific to the dataset or the data distribution it works on. However, our initial assumption was that the pipeline might be more or less independent from the algorithm. Therefore, we would like to quantify how much an optimal configuration is specific to an algorithm or is {\it universal}, i.e. works well regardless of the algorithm. For this, the optimization process might be performed on a collection of methods $\mathcal{A} = \{ A_i \}^N_{i=1}$. The result is a sample of optimal configurations $\mathbf{p}^* = \{ p^*_{i} \}^M_{i=1}$ where $M \geq N$ since an algorithm might have several distinct optimal configurations. After normalizing the configuration space to bring each axis to $[0,1]$, the link between the processed data and the methods can be studied through a new indicator named Normalized Mean Absolute Deviation (NMAD). The idea behind this metric is to measure how much the optimal points are distant from a reference optimal point. If the optimal configuration does not depend on the algorithm, the expected distance between the optimal configurations is 0. Conversely, if a point is specific to an algorithm, the other points will be in average far from it.

Working in the normalized configuration space has two advantages. First, it forces all parameters to have the same impact. Secondly, it allows the comparison from one dataset to another since the NMAD belongs to $[0,1]$ for any number of algorithms or dimensions of the configuration space.


The Normalized Mean Absolute Deviation is the norm 1 of the Mean Absolute Deviation\footnote{As we work on a discrete space, we used the norm 1, but the Euclidean norm is probably a better choice in continuous space.}, divided by the number of dimensions $K$ of the configuration space.
\begin{definition}[Normalized Mean Absolute Deviation (NMAD)]
  \begin{displaymath}
    \text{NMAD}(\mathbf{p}^*, r) = \frac 1 K  \frac 1 N || \big( \underset{i = 1}{\overset{N}{\sum}} |p^*_{i} - r|\big) ||_1
  \end{displaymath}
\end{definition}

To measure how much each optimal point $p^*_i$ is specific to an algorithm $A_j$, we use it as a reference point and calculate the NMAD using a sample composed of all the optimal points. However, an algorithm might have several optimal points and to be fair, we use as a representant of each algorithm, the closest point to the reference point.

\subsection{Experimental Settings}


As the configuration space described in Section \ref{sec:protocol_1} is not a metric space, we cannot directly use the NMAD. To avoid introducing bias with an {\it ad-hoc} distance, we perform another experiment with a configuration space that is embedded in $\mathbb{N}$.

We collected 1000 judgements documents provided by the European Court of Human Rights (ECHR) about the Article 6. The database HUDOC\footnote{\url{https://hudoc.echr.coe.int/}} provides the ground truth corresponding to a violation or no violation. The cases have been collected such that the dataset is balanced. The conclusion part is removed.
To confirm the results, we used a second dataset composed of 855 documents from the categories atheism and religion of 20newsgroups.

Each document is preprocessed using a data pipeline consisting in tokenization, stopwords removal, followed by a $n$-gram generation. The processed documents are combined and the $k$ most frequent tokens across the corpus are kept, forming the dictionary. Each case is turned into a Bag-of-Words using the dictionary.

There are two hyperparameters in the preprocessing phase: $n$ the size of the $n$-grams, and $k$ the number of tokens in the dictionary. We defined the parameter configuration domain as follows:
\begin{itemize}
\item $n \in \{1,2,3,4,5\}$,
\item $k \in \{10, 100, 1000, 5000, 10000, 50000, 100000\}$.
\end{itemize} 
We used the same four algorithms as in Section \ref{sec:exp_1}. As we are interested in the optimal configurations, we performed an exhaustive search.

\subsection{Results}
\label{sec:r}


\begin{figure}
\includegraphics[scale=0.4]{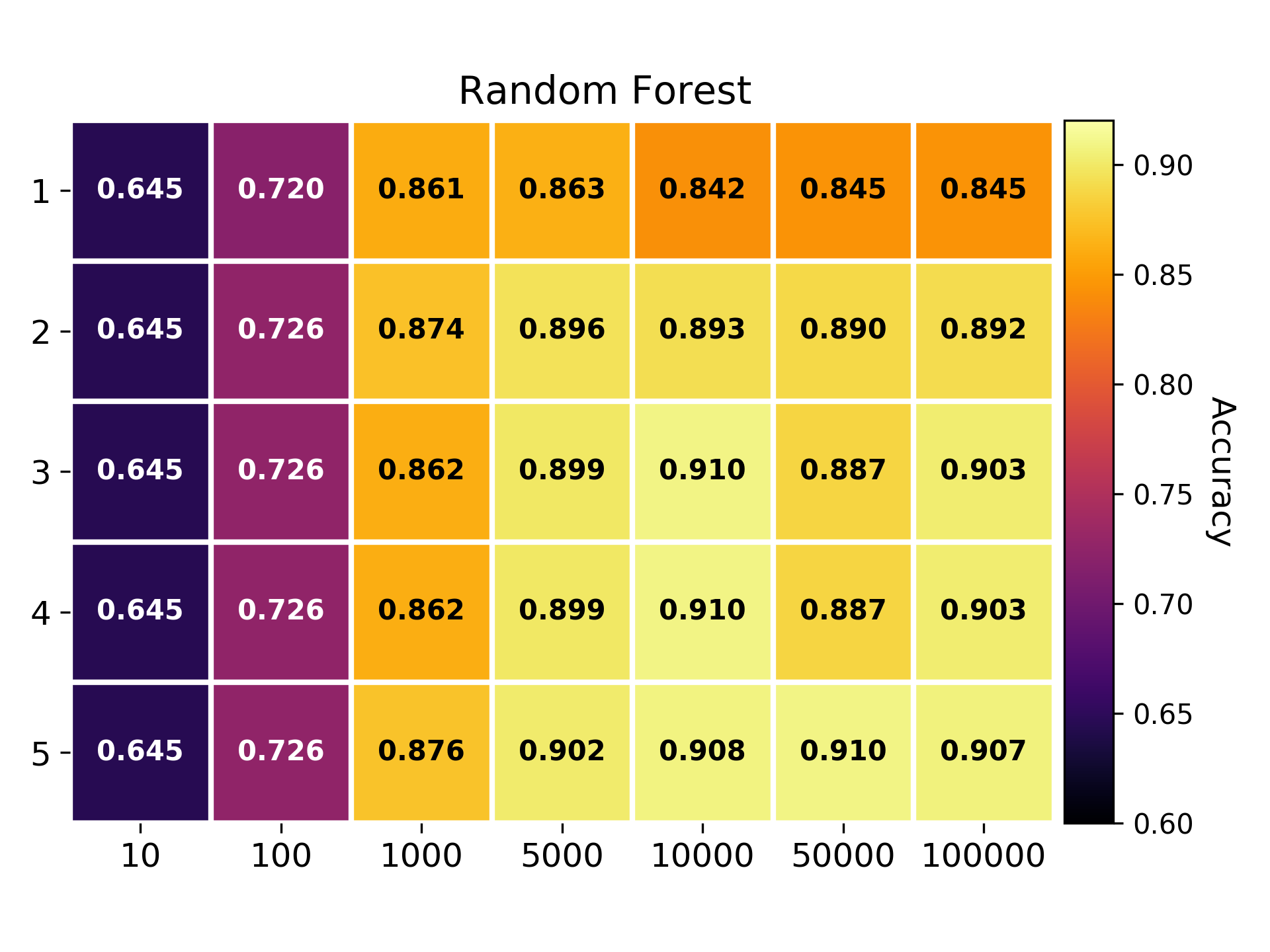}\hfill
\includegraphics[scale=0.4]{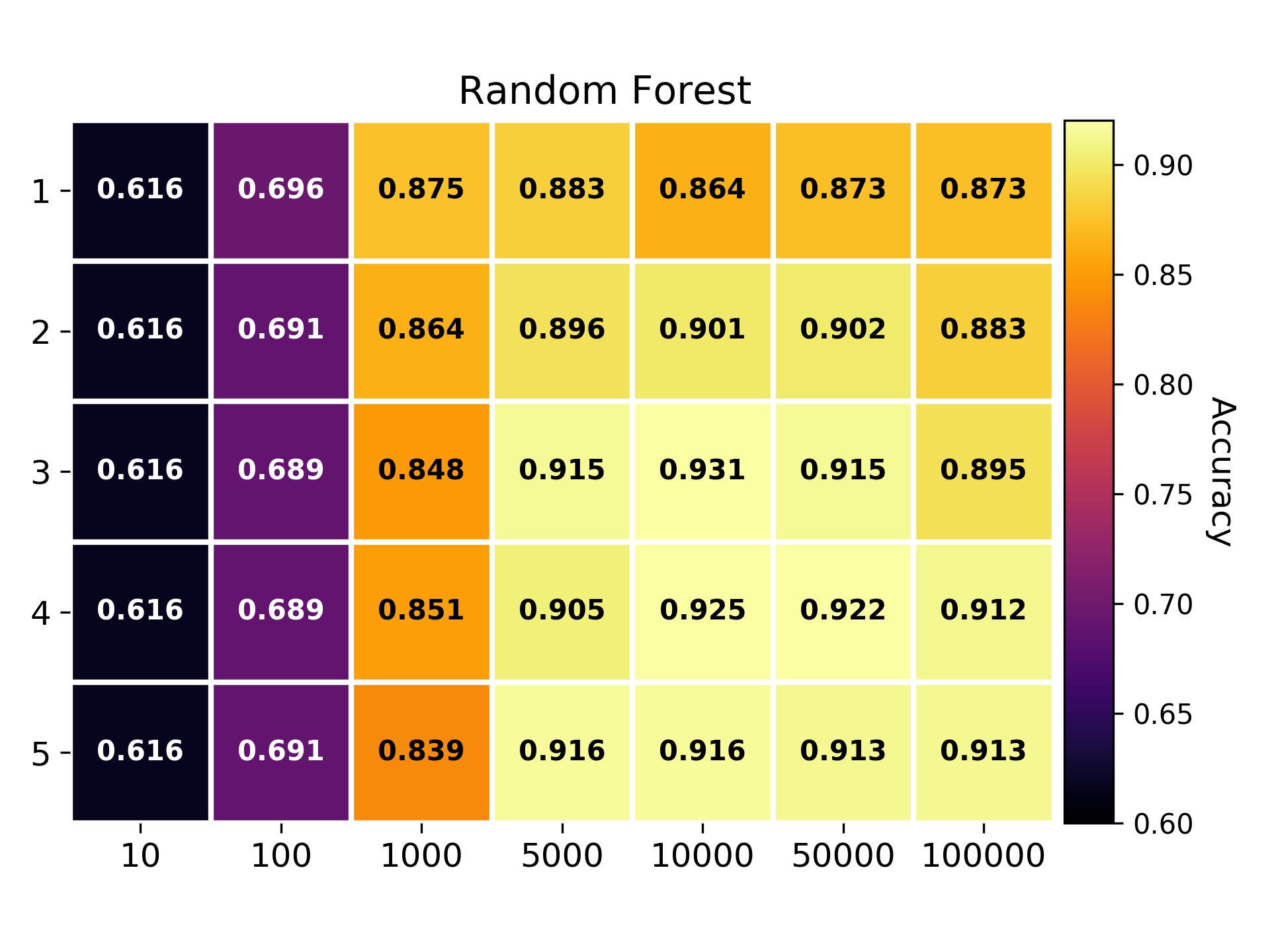}
\caption{Heatmap depicting the accuracy depending on the pipeline parameter configuration on ECHR (left) and Newsgroup (right).}
\label{heatmap}
\end{figure}

For both datasets, Figure \ref{heatmap} shows that the classifier returns poor results for a configuration with a dictionary of only 10 or 100 tokens. Both parameters influence the results, and too high values deteriorate the results.

Table \ref{tab:best_config} summarizes the best configurations per method. For the first dataset, there are 3 points that gives the optimal value for Random Forest and Linear SVM, however, in practice lowest parameters values are better because they imply a lower preprocessing and training time. It is interesting to notice that $(5, 50000)$ returns the best accuracy for every model, as this point would be a sort of {\it universal} configuration for the dataset, taking the best out of the data source, rather than being well suited for a specific algorithm. On the contrary, on Newsgroup, all optimal points are different. Our hypothesis is that the more structured a corpus is, the less algorithm-specific are the optimal configurations, because the preprocessing steps become more important to extract markers used by the algorithms to reach good performances. As ECHR dataset describes standardized justice documents, it is far more structured than Newsgroup. This would also explain why generating $n$-grams for $n=5$ still improves the results on ECHR while degrading them on Newsgroup.

\begin{table}
  \center
  \caption{Best configurations depending on the method}
  \label{tab:best_config}

  \begin{tabular}{rrl}
    \toprule
    Method & $(n,k)$ & accuracy\\
    \midrule
    \multicolumn{3}{c}{ECHR}\\
    \midrule
    Decision Tree &(5, 50000) &0.900\\
    Neural Network & (5, 50000) & 0.960 \\
    Random Forest &(3, 10000), (4, 10000), (5, 50000)& 0.910\\
    Linear SVM & (3, 50000), (4, 50000), (5, 50000)& 0.921\\
    \midrule
    \multicolumn{3}{c}{Newsgroup}\\
    \midrule
    Decision Tree & (4, 5000), (4, 100000) & 0.889 \\
    Neural Network & (5, 50000) & 0.953 \\
    Random Forest & (3, 10000) & 0.931\\
    Linear SVM & (2, 100000) & 0.946\\
  \bottomrule
\end{tabular}
\end{table}


This hypothesis is partially confirmed by Table \ref{tab:param_m}, where it is clear that the $n$-gram operator has a strong impact on the accuracy variation on ECHR dataset (up to 9.8\% accuracy improvement) while almost none on Newsgroup dataset (at the exception of Random Forest). 

\begin{table}[h!]
  \caption{Impact of parameter $n$ on the accuracy, measured as the relative difference between the best results obtained only using $(1, k)$ and the best results obtained for any configuration $(n,k)$.}
  \label{tab:param_m}
\center
  \begin{tabular}{rccc}
    \toprule
    Method & $p=(1, k)$ & $p=(n,k)$ & $\Delta$ acc\\
    \midrule
    \multicolumn{4}{c}{ECHR}\\
    \midrule
    Decision Tree & 0.850 & 0.900 & 5.9\%\\
    Neural Network & 0.874 & 0.960 & 9.8\%\\
    Random Forest & 0.863 & 0.910 & 5.4\%\\
    Linear SVM & 0.892 & 0.921 & 6.6\%\\
    \midrule
    \multicolumn{4}{c}{Newsgroup}\\
    \midrule
    Decision Tree &0.885 & 0.889 & 0.5\%\\
    Neural Network &0.949 & 0.953 & 0.4\%\\
    Random Forest & 0.883 & 0.931 & 5.4\% \\
    Linear SVM & 0.945 & 0.946 & 0.1\%\\
  \bottomrule
\end{tabular}
\end{table}
\noindent
Table \ref{tab:nmad} contains the NMAD value for each distinct optimal configuration reported in Table \ref{tab:best_config}. 
As expected, the point $(5, 50000)$ has a NMAD of 0 since the point is present for every algorithm: $(5, 50000)$ is a {\it universal} pipeline configuration for this data pipeline and dataset. The point $(4, 50000)$ appears only once but it is really close to $(5, 50000)$ (itself in the 3 other algorithms results) s.t. its NMAD is low. It can be interpreted as belonging to the same area of optimal values. On the opposite, $(3, 10000)$ and $(4, 10000)$ have high NMAD w.r.t. the other points, indicating they are isolated points and may be algorithm specific. Their NMAD values are rather low because despite the points are isolated, they differ significantly from the others points only on the second component. In comparison, if $(1, 10)$ would be an optimal point for Random Forest, its NMAD would be 0.5. 
On the contrary, for Newsgroup, the NMAD value is rather high and similar for all points, indicating that they are at a similar distance from each other and really algorithm specific. 

To summarize, the NMAD metric is coherent with the conclusion drawn from the heatmaps and Table \ref{tab:best_config}, and suggests that there exist two types of optimal configurations: {\it universal} pipeline configurations that work well on a large range of algorithms for a given dataset, and algorithm-specific configurations. Thus, we are confident the NMAD can be used in larger configuration spaces where heatmaps and exhaustive results are not available for graphical interpretation, and help to reuse configurations and initialize surrogate models.

\begin{table}
  \caption{Normalized Mean Average Deviation for each optimal configuration found.}
  \label{tab:nmad}
  \center
  \begin{tabular}{lrl}
    \toprule
    \multicolumn{2}{c}{ECHR}\\
    \midrule
     Point & NMAD\\
    \midrule
     (5, 50000) & 0\\
     (3, 10000) & 0.275\\
     (4, 10000) & 0.213\\
     (3, 50000) & 0.175\\
     (4, 50000) & 0.094\\
  \bottomrule
\end{tabular}\qquad
\begin{tabular}{lrl}
    \toprule
    \multicolumn{2}{c}{Newsgroup}\\
    \midrule
     Point & NMAD\\
    \midrule
    (4, 5000) & 0.306\\
    (4, 100000) & 0.300 \\
    (5, 50000) & 0.356 \\
    (3, 10000) & 0.294 \\
    (2, 100000)& 0.362 \\
  \bottomrule
\end{tabular}
\end{table}



\section{Conclusion}
\label{sec:c}

In this paper, we proposed a reformulation of the main \automl~problem to fit better the way data scientists work in practice. We studied the importance of the data pipeline and algorithm hyperparameter tuning phase on a reasonably realistic search space. We found that spending more time on setting a proper data pipeline usually leads to better results. We proposed a two-stage process to optimize a general machine learning workflow, articulated around the data pipeline construction and the hyperparameter tuning. Under time constraint, we studied four different policies to allocate time between these two phases. We found that the best results are obtained with a tradeoff that seems to depend on both the algorithm and the dataset. Last, we found that the most robust policies, in terms of reaching a good score within little time, are the most advanced policies: iterative and adaptive.

In addition, we provided a metric to study if a data pipeline is more or less independent from the algorithm. This metric can be use to suggest good data pipelines depending on meta-features as described in \cite{BILALLI2018101,Bilalli:2017:PPM:3214035.3214049}]. Future work should focus on an online version s.t. the pipeline is tuned in a streaming way. Also, the NMAP indicator works only in euclidian spaces which is not the case for the first experiment. Therefore, further work should focus on extending the NMAP to non-vector space.

More generally, we plan a larger experimental campaign using \OpenML~benchmark suit \cite{bischl2017openml} with different pipeline prototypes.

A generalization to multi-stage optimization is also considered. Each layer of the pipeline could benefit from a specific time allocation depending on its marginal contribution. Some preliminary work has been done toward this direction in \cite{DBLP:journals/corr/abs-1903-02521}.

Last but not least, we are currently working on a library based on the code of those experiments, dedicated to the creation of flexible pipelines and budget allocation strategies. By using the operator and pipeline prototype definitions provided in this paper, we will provide an easy yet generic way to define and extend pipeline search spaces, which is currently not feasible by any \automl~systems.

\bibliography{bibliography}
\newpage
\appendix

\section{Pipeline configuration space}

\begin{table}[h]

\caption{Pipeline search space.}
\label{table:pipeline_search_space}

\centering

\begin{tabular}{lcccc}
\toprule
& $\# \lambda$ & $|\Lambda|$ & impl. & \\
\midrule
\multicolumn{4}{l}{\bf Rebalance}\\
\midrule
No operator & 0 & 0 & - \\
Near Miss & 1 & 3 & \vimblearn \\
Condensed Nearest Neighbour & 1 & 3 & \vimblearn \\
SMOTE & 1 & 3 & \vimblearn \\
\midrule
\multicolumn{4}{l}{\bf Normalize}\\
\midrule
No operator & 0 & 0 & - \\
Standard Scaler & 2 & 4 & \vsklearn \\
Power Transform & 0 & 0 & \vsklearn \\
MinMax Scaler & 0 & 0 & \vsklearn  \\
Robust Scaler & 3 & 12 &  \vsklearn  \\
\midrule
\multicolumn{4}{l}{\bf Features}\\
\midrule
No operator & 0 & 0 & - \\
PCA & 1 & 4 &  \vsklearn \\
SelectKBest (F-score) & 1 & 4 &  \vsklearn \\
PCA $\cup$ SelectKBest & 2 & 16 & \vsklearn \\
\bottomrule
\end{tabular}

\begin{flushleft} \small

The column $\# \lambda$ is the number of parameters while $|\Lambda|$ is the total number of values in the operator configuration space. The column {\it impl.} indicates the implementation of the operator (scikit-learn or imbalanced-learn).
\end{flushleft}
\end{table}









\end{document}